\documentclass[sigconf, screen]{acmart}
\renewcommand\footnotetextcopyrightpermission[1]{}
\AtBeginDocument{%
  }

\setcopyright{acmlicensed}
\copyrightyear{2018}
\acmYear{2018}
\acmDOI{XXXXXXX.XXXXXXX}
\acmConference[Conference acronym 'XX]{Make sure to enter the correct
  conference title from your rights confirmation email}{June 03--05,
  2018}{Woodstock, NY}
\acmISBN{978-1-4503-XXXX-X/2018/06}

\usepackage{booktabs}
\usepackage[table]{xcolor}
\usepackage{subcaption}
\usepackage{graphicx} 

\usepackage{amsmath}
\usepackage{mathtools}
\usepackage{amsthm}

\usepackage{booktabs}
\usepackage{multirow}
\usepackage{makecell}
\usepackage{array}
\usepackage[export]{adjustbox}
\usepackage[capitalize,noabbrev]{cleveref}
\usepackage[most]{tcolorbox}
\usepackage{subcaption}
\newtcolorbox{algopanel}[1]{%
  enhanced,
  colback=black!6,      
  colframe=black,
  boxrule=0.8pt,
  sharp corners,
  left=2mm,right=2mm,
  top=7mm,bottom=2mm,
  before skip=6pt, after skip=6pt,
  overlay={%
    \fill[black!80] (frame.north west) rectangle ([yshift=-7mm]frame.north east);
    \node[
      anchor=west,
      font=\bfseries\footnotesize,
      text=white
    ] at ([xshift=2mm,yshift=-3.5mm]frame.north west) {#1};
  },
}
\usepackage{tabularx}
\usepackage{float}

\begin{document}

\title{MaLoRA: Gated Modality LoRA for Key-Space Alignment in Multimodal LLM Fine-Tuning}

\author{Xinhan Zheng}
\authornote{Both authors contributed equally to this research.}
\email{xinhanzheng@mail.ustc.edu.cn}
\affiliation{%
  \institution{University of Science and Technology of China}
  \city{Hefei}
  \country{China}}

\author{Huyu Wu}
\authornotemark[1]
\email{wuhuyu25@mails.ucas.ac.cn}
\affiliation{%
  \institution{University of Chinese Academy of Sciences}
  \city{Beijing}
  \country{China}}

\author{Xueting Wang}
\email{wangxueting@mail.ustc.edu.cn}
\authornotemark[1]
\affiliation{%
  \institution{University of Science and Technology of China}
  \city{Hefei}
  \country{China}}

\author{Duo Su}
\email{suduo@mail.tsinghua.edu.cn}
\affiliation{%
  \institution{Tsinghua University}
  \city{Beijing}
  \country{China}}

\author{Haiyun Jiang}
\authornote{Corresponding author.}
\email{haiyunjiang@sjtu.edu.cn}
\affiliation{%
  \institution{Shanghai Jiao Tong University}
  \city{Shanghai}
  \country{China}}

\renewcommand{\shortauthors}{Zheng et al.}

\begin{abstract}
Multimodal large language models (MLLMs) often exhibit \emph{text-centric bias} under joint image--text inputs, over-relying on textual signals and under-using visual evidence. We analyze decoder self-attention and observe a persistent cross-modal misalignment in the attention \emph{key space}, where visual and text keys form separated distributions consistent with attention favoring text tokens. Motivated by this finding, we propose \textbf{Modality Alignment LoRA (MaLoRA)}, a fine-tuning framework that targets key-space misalignment via three designs: Gated Modality LoRA (GML) for modality-conditioned key adaptation, multi-kernel maximum mean discrepancy (MMD) for cross-modal distribution alignment, and Gram reference regularization to preserve within-modality structure during alignment.  Extensive experiments across three MLLM backbones and diverse benchmarks demonstrate that MaLoRA reduces key-space divergence and yields measurable improvements in downstream performance.
\end{abstract}




\received{20 February 2007}
\received[revised]{12 March 2009}
\received[accepted]{5 June 2009}

\maketitle

\section{Introduction}
Multimodal large language models (MLLMs) have demonstrated strong performance across vision–language tasks by integrating visual and textual signals within a unified generative framework. Ideally, such models should dynamically balance modalities according to task-relevant evidence. However, empirical observations reveal a systematic deviation: under joint image–text inputs, MLLMs exhibit a pronounced preference for textual information, often neglecting visual cues even when they are essential for correct reasoning. 
This \textbf{text-centric bias} persists across architectures, datasets, and training regimes, suggesting that it reflects a fundamental property of multimodal generative modeling rather than a superficial artifact of data or optimization.

Existing approaches primarily address this issue through data curation, alignment objectives, architectural modifications, or inference-time heuristics~\cite{chen2024sharegpt4v,sun2024aligning,huang2024opera,leng2024mitigating,yin2024woodpecker}. However, the model-internal mechanism underlying text-dominant behavior remains insufficiently understood. This gap motivates us to investigate how modality imbalance emerges within the attention computation of MLLMs.

In this work, we argue that text-centric bias originates from a structural property of decoder self-attention in MLLMs: \textbf{cross-modal misalignment in the attention key space}. 
Formally, although visual and textual tokens are projected through shared key matrices, their resulting key representations follow distinct distributions. 

Since decoder queries are shaped predominantly by language-model pretraining, they align more naturally with textual key distributions, yielding systematically higher similarity scores with text tokens than visual. 
This induces a biased attention allocation mechanism that is intrinsic to the geometry of the key space rather than the semantics of the input.

To examine the hypothesis that text-centric bias in MLLMs is linked to cross-modal misalignment in the attention key space, we conduct a multi-layer analysis of key representations in several representative MLLMs.
By visualizing key distributions and quantifying their divergence using distributional metrics, we demonstrate that visual and textual keys form persistently separated manifolds across layers and models. Moreover, the magnitude of cross-modal divergence significantly exceeds within-modality variation, establishing key-space misalignment as a persistent and fundamental phenomenon. These observations suggest that modality bias is not merely an optimization artifact but a consequence of distributional mismatch in the latent space that governs attention computation.

This perspective reframes multimodal bias mitigation as a  \textit{distribution alignment problem in the attention key space}. However, directly aligning visual and textual keys raises two theoretical challenges. First, visual and textual tokens require modality-specific adaptation directions, making shared parameter updates insufficient and potentially conflicting. Second, naive alignment risks collapsing or distorting the intrinsic geometry of modality-specific representations, thereby degrading the  reasoning capacity of models. Therefore, an effective intervention must simultaneously enable modality-conditioned adaptation and preserve structural properties of the original representation space.

To address these challenges, we propose \textbf{MaLoRA}, a fine-tuning framework that targets key-space misalignment. MaLoRA introduces (i) \textbf{Gated Modality LoRA}, which decomposes key adaptation into modality-specific low-rank subspaces, (ii) \textbf{distribution-level alignment} via multi-kernel maximum mean discrepancy to reduce cross-modal divergence, and (iii) \textbf{structure-preserving regularization} based on Gram reference to constrain geometric drift during alignment. Together, these components provide a principled mechanism for reshaping the key-space geometry while maintaining representational integrity, without altering the inference process.

Extensive experiments across multiple MLLM backbones and benchmarks show that MaLoRA consistently reduces cross-modal key divergence and improves performance, particularly in tasks that require genuine visual grounding. Beyond empirical gains, our analysis highlights the attention key space as a previously underexplored locus of multimodal bias and offers a theoretically grounded framework for understanding and mitigating modality imbalance in generative multimodal models.

Our contributions are as follows:
\begin{itemize}
\item Starting from standard self-attention, we connect \emph{text-centric bias} to the attention \emph{key space} and identify cross-modal key misalignment as a key contributing factor. We hypothesize that cross-modal key distribution misalignment correlates with a systematic preference for text tokens, and support it with multi-layer visualization and divergence measurements across backbones and benchmarks.
\item We propose \textbf{MaLoRA}, a fine-tuning framework that directly intervenes on key-space misalignment: \textbf{Gated Modality LoRA (GML)} enables modality-conditioned key adaptation, and \textbf{MMD} alignment together with \textbf{Gram reference} structure preservation reduces cross-modal gaps while avoiding excessive distortion of representation geometry.
\item We validate MaLoRA on evaluations spanning diverse tasks and datasets. The method consistently reduces distribution divergence between visual and textual representations and improves downstream performance.
\end{itemize}

\begin{figure*}[t]
    \centering
    \includegraphics[width=0.88\linewidth]{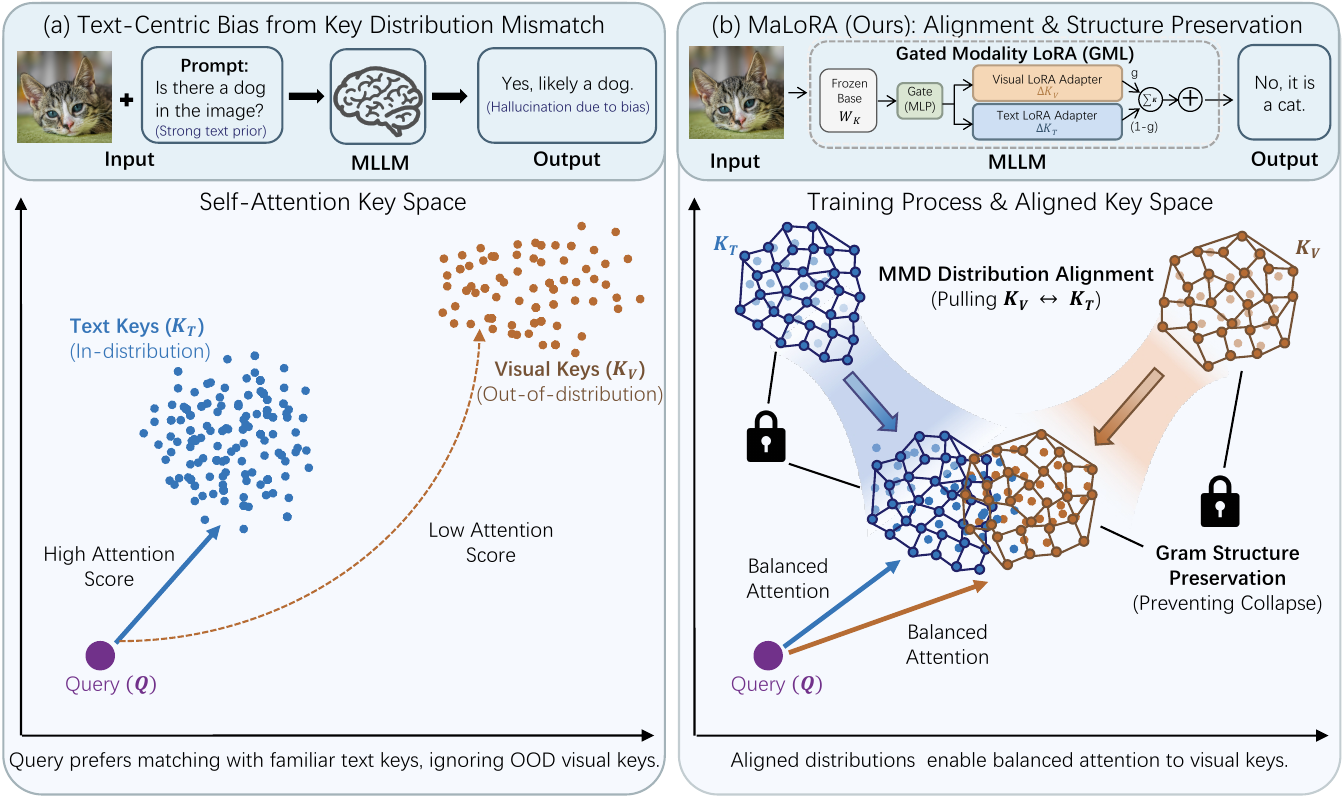}
    \caption{The MaLoRA framework addressing \emph{text-centric bias} in MLLMs.}
    \label{fig:MaLoRA_overview}
\end{figure*}
\section{Related Work}
\label{sec:related}

\subsection{Modality bias and text-centric bias in MLLMs}
MLLMs often exhibit modality imbalance in generative tasks: they rely on linguistic priors over visual evidence and may generate hallucinated content when visual cues are weak or not aligned with question-critical evidence~\cite{huang2024opera,leng2024mitigating,yin2024woodpecker,sun2024aligning}. Recent work has begun to systematically characterize this \emph{text dominance} phenomenon. \citet{wu2025language} analyze its prevalence across multiple non-text modalities (images, videos, audio) and proposes quantitative metrics to assess the degree of modality dominance. Meanwhile, \citet{park2025assessing} argues that benchmarks contain limited cases that truly require multimodal fusion, which amplifies unimodal bias. \citet{liu2025modality} discusses modality imbalance under alignment and preference optimization and proposes optimization frameworks for modality balance. Complementary to these perspectives on phenomenon, data, and training strategy, we further analyze the internal mechanism of text dominance from the self-attention \emph{key space} distribution, and show that the out-of-distribution nature of visual keys can systematically affect attention allocation and generation behavior.
\subsection{Structural or inference-time mitigation methods}
Existing mitigation strategies mainly operate either by modifying multimodal fusion and training signals or by suppressing hallucination at inference time.
 
To address weakened visual evidence usage and hallucinations caused by text dominance, one line of work modifies fusion architectures or training signals to increase the contribution of visual information in generation. For example, LACING introduces dual-branch attention for vision and text at the attention layer, combined with soft visual prompts to reduce text-centric bias~\citep{zhao2025lacing}. However, LACING operates in the full-parameter multimodal alignment setting, requiring complete two-stage retraining of the model from the LLM backbone (e.g., 558K pretraining + 665K instruction tuning on 8×A100 GPUs), and further relies on a contrastive decoding strategy at inference time. In contrast, MaLoRA targets the parameter-efficient fine-tuning regime, adapting pre-trained instruct models to downstream tasks with only lightweight LoRA updates and standard autoregressive decoding. The two approaches are thus complementary, addressing text-centric bias at different stages of the MLLM lifecycle.
Unlike MoE architectures that primarily expand capacity via dynamic routing, GML introduces modality-specific LoRA branches within $k_{proj}$ as separate adaptation interfaces for visual and textual tokens.

Another line of work keeps model parameters fixed and suppresses erroneous tokens triggered by linguistic priors at inference time via contrastive decoding or probabilistic penalties. VCD reduces language priors by contrasting output distributions between the original image and a perturbed image~\citep{leng2024mitigating}. OPERA applies an over-trust penalty in beam search based on aggregated attention patterns and performs backtracking to re-rank candidates~\citep{huang2024opera}. ICD contrasts the distributions under standard instructions and ``perturbed instructions'' to offset hallucinated concepts~\citep{wang2024mitigating}. CATCH further adopts adaptive token-level contrast to mitigate accumulated hallucinations in open-ended generation~\citep{kan2024catch}. 

Overall, these methods intervene at the architecture or decoding stage. In contrast, training-time constraints that directly target cross-modal discrepancy in the decoder self-attention \emph{key space} remain relatively underexplored. In many cases, inference-time methods introduce extra decoding steps or additional forward passes, and they do not directly alter representation misalignment. We instead intervene during fine-tuning by constraining cross-modal gaps in the attention \emph{key space}, while keeping inference unchanged.

\subsection{Cross-modal alignment and distribution alignment}
Cross-modal representation alignment and distribution matching are common approaches in multimodal learning and domain generalization/adaptation. They include optimal-transport (OT) based structure-preserving alignment \citep{chen2025prompt} and kernel-based statistical distances such as maximum mean discrepancy (MMD) \citep{gretton2012kernel,sun2024craft}. In addition, \citet{song2024set} studies semi-supervised alignment from a distributional perspective and discusses training paradigms that improve cross-modal semantic consistency under weak pairing or limited alignment data. Existing distribution alignment work mainly targets representation learning and adaptation settings. We introduce distribution alignment into the attention key space of MLLMs to address their modality-specific bias mechanism.

\section{Key-Space Misalignment}
\label{sec:key_space_misalignment}
Under image–text inputs, MLLMs often over-rely on textual information during generation, exhibiting \emph{text-centric bias}.
\textbf{\textit{We hypothesize that this phenomenon is not solely due to data composition or alignment objectives, but may stem from \emph{key-space distribution misalignment} in decoder self-attention.}}
We recall the standard self-attention formulation:
\begin{equation}
\mathrm{Attn}(\mathbf{Q},\mathbf{K},\mathbf{V})
=
\mathrm{softmax}\!\left(\frac{\mathbf{Q}\mathbf{K}^{\top}}{\sqrt{d}}\right)\mathbf{V},
\label{eq:attn}
\end{equation}
where $\mathbf{Q}$, $\mathbf{K}$, and $\mathbf{V}$ denote the query, key, and value matrices, respectively, and $d$ is the per-head dimension. At the token level, the attention score between a query $\mathbf{q}_i$ and a key $\mathbf{k}_j$ is determined by their scaled dot product $\mathbf{q}_i^\top \mathbf{k}_j / \sqrt{d}$.
In contrast, visual tokens are projected from a vision encoder and concatenated with text tokens, inheriting a cross-modal gap before entering the decoder. At each layer, both visual and textual keys are produced from their hidden states through a shared key projection $\mathbf{W}_k$.
Despite sharing the projection, $\mathbf{K}_{\text{vis}}$ and $\mathbf{K}_{\text{text}}$ may still exhibit substantial distribution differences in the key space. 
As a result, $\mathbf{q}$ may be better aligned with $\mathbf{k}^{\text{text}}$, yielding higher similarity scores and attention allocation bias toward text tokens.

To test this hypothesis, we perform \emph{key-space probing}, extracting keys and analyzing their geometry and divergence, on several decoder layers of representative MLLMs.
Given joint image-text inputs, we extract the sets of visual and textual key vectors${\mathbf{k}^{\text{vis}}}, {\mathbf{k}^{\text{text}}}$ from target layers and mask padding tokens.
First, we visualize the geometry of the two key sets on selected layers using t-SNE.
Second, we quantify cross-modal discrepancy using distributional measures (primarily MMD), comparing $P(\mathbf{K}^{\text{vis}})$ and $P(\mathbf{K}^{\text{text}})$ and contrasting them with within-modality baselines computed from split subsets.

\begin{figure*}[t]
  \centering
  \scalebox{0.9}{%
  \begin{minipage}[c]{0.55\textwidth}
    \centering
    \begin{tabular}{
      @{}
      c
      @{\hspace{2pt}}
      c
      @{\hspace{6pt}}
      c
      @{}
    }
      &
      \makebox[0.42\textwidth]{\small\textbf{LLaVA-1.5-7B on MMMU}}
      &
      \makebox[0.42\textwidth]{\small\textbf{LLaVA-1.5-7B on MMBench-EN}}
      \\[2pt]
      \raisebox{-.5\height}{\rotatebox{90}{\small\textbf{Base}}}
      &
      \raisebox{-.5\height}{\includegraphics[width=0.42\textwidth]{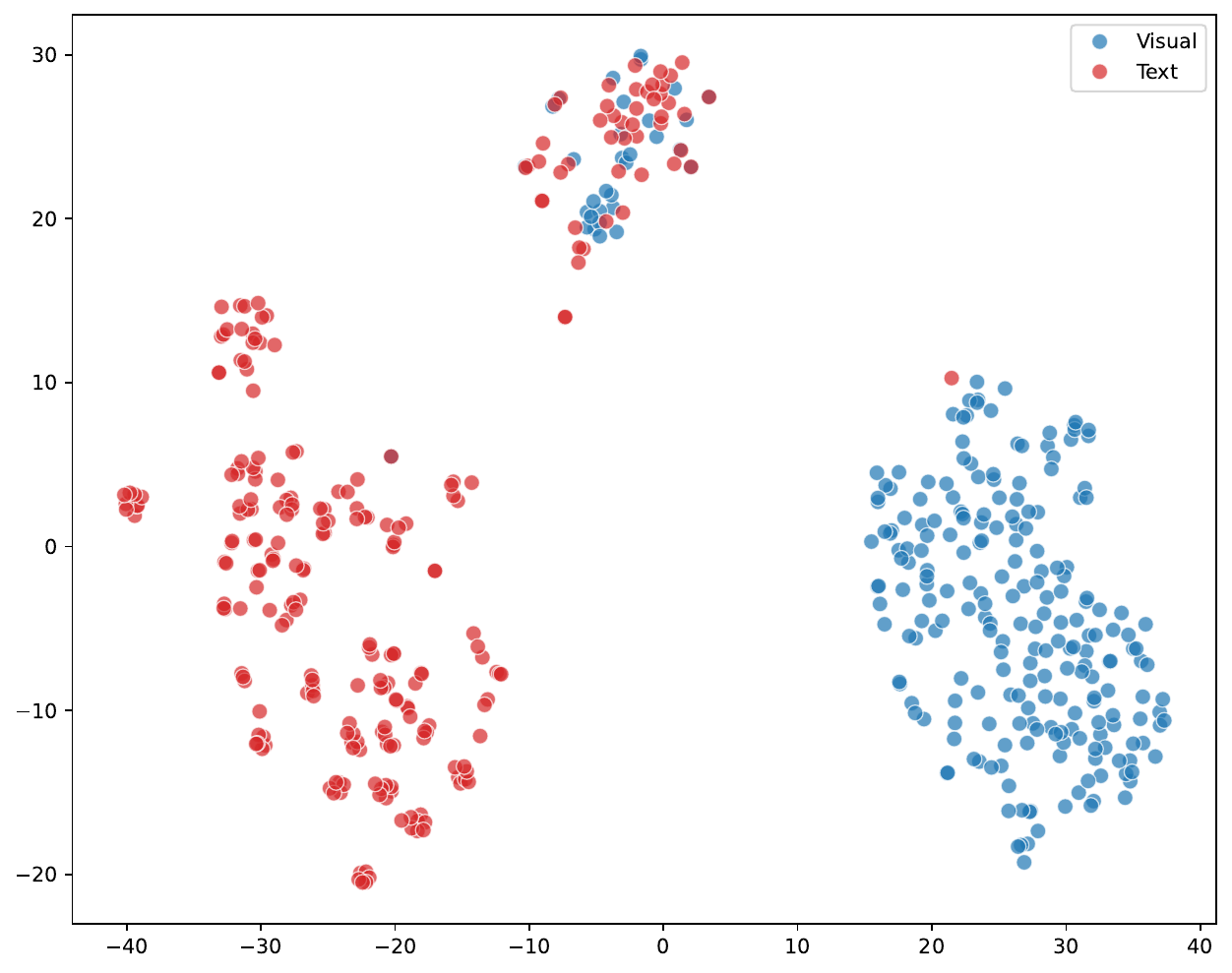}}
      &
      \raisebox{-.5\height}{\includegraphics[width=0.42\textwidth]{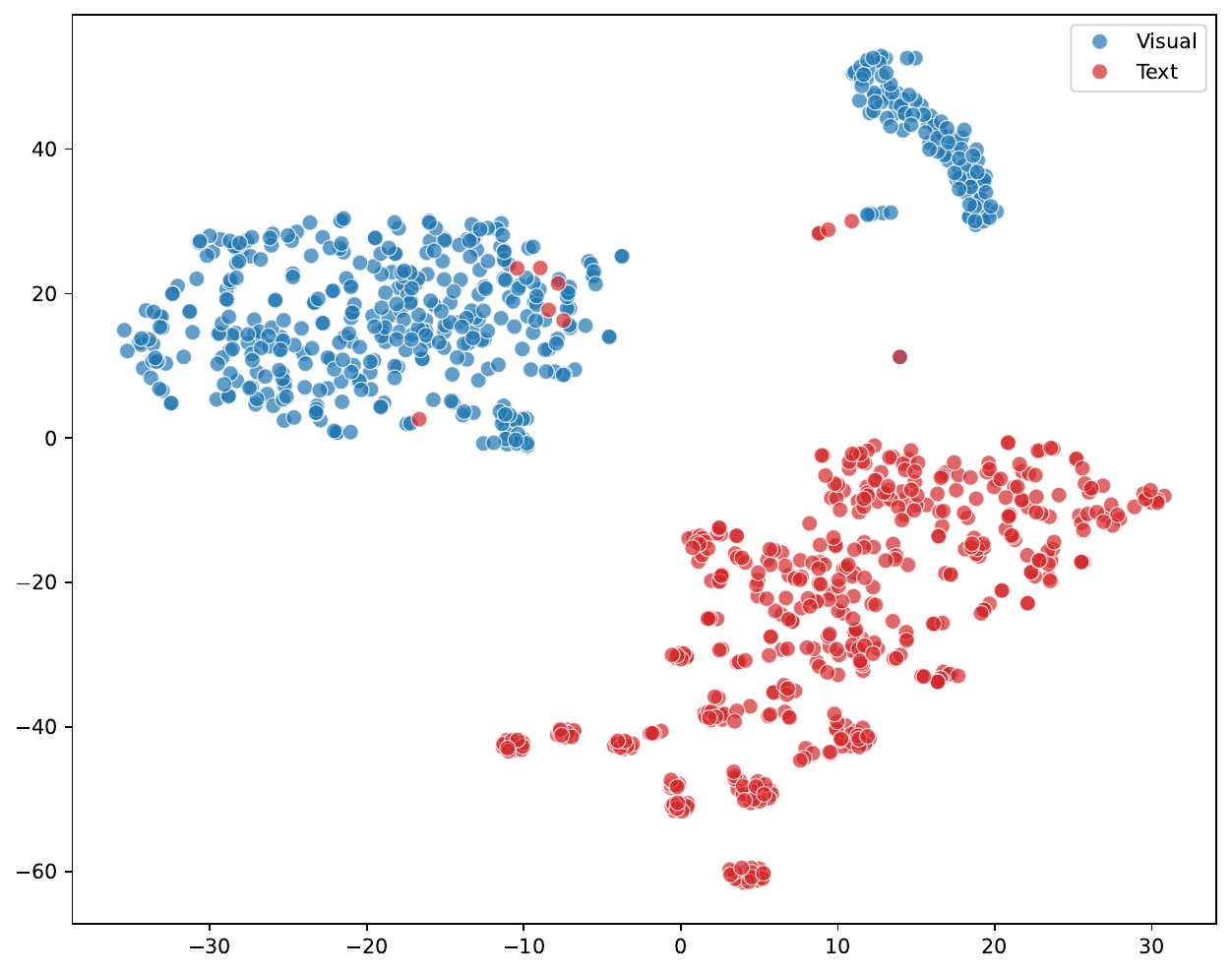}}
      \\[4pt]
      \raisebox{-.5\height}{\rotatebox{90}{\small\textbf{MaLoRA}}}
      &
      \raisebox{-.5\height}{\includegraphics[width=0.42\textwidth]{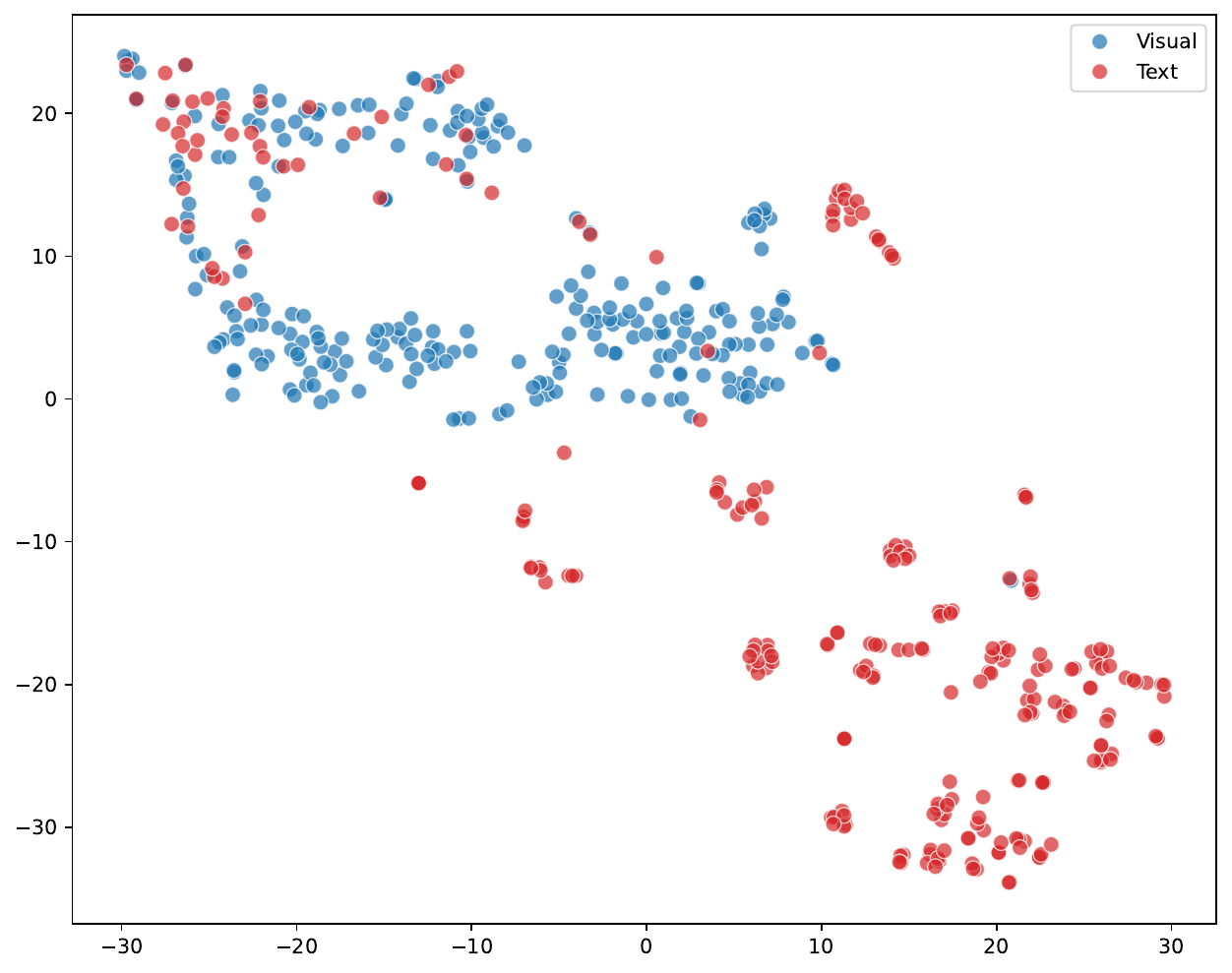}}
      &
      \raisebox{-.5\height}{\includegraphics[width=0.42\textwidth]{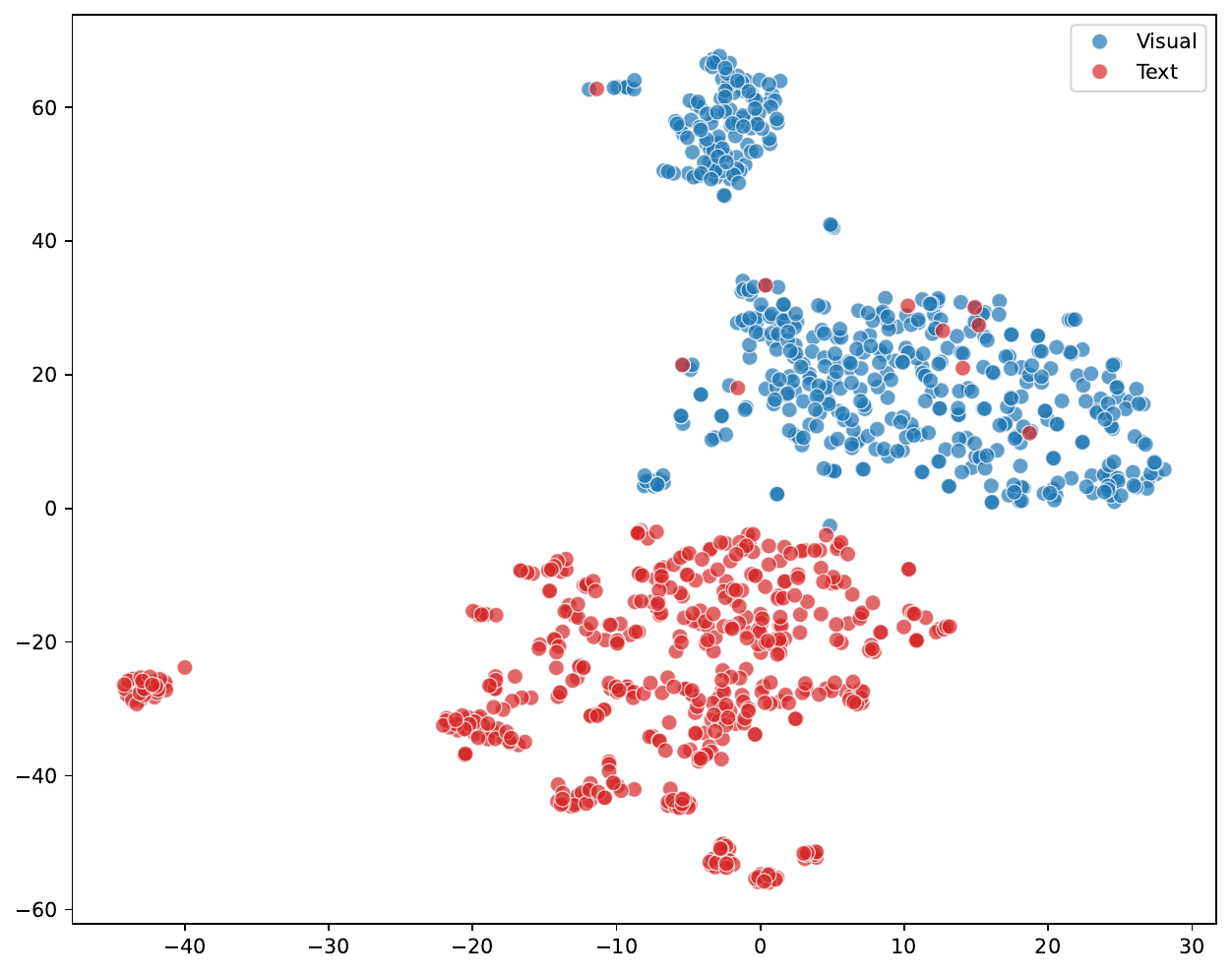}}
      \\
    \end{tabular}
  \end{minipage}%
  \hfill
  \begin{minipage}[c]{0.40\textwidth}
    \centering
    \begin{tabular}{@{}c@{}}
      \makebox[\linewidth]{\small\textbf{Qwen3-VL-8B on MMBench-EN}} \\
      \includegraphics[width=0.92\linewidth]{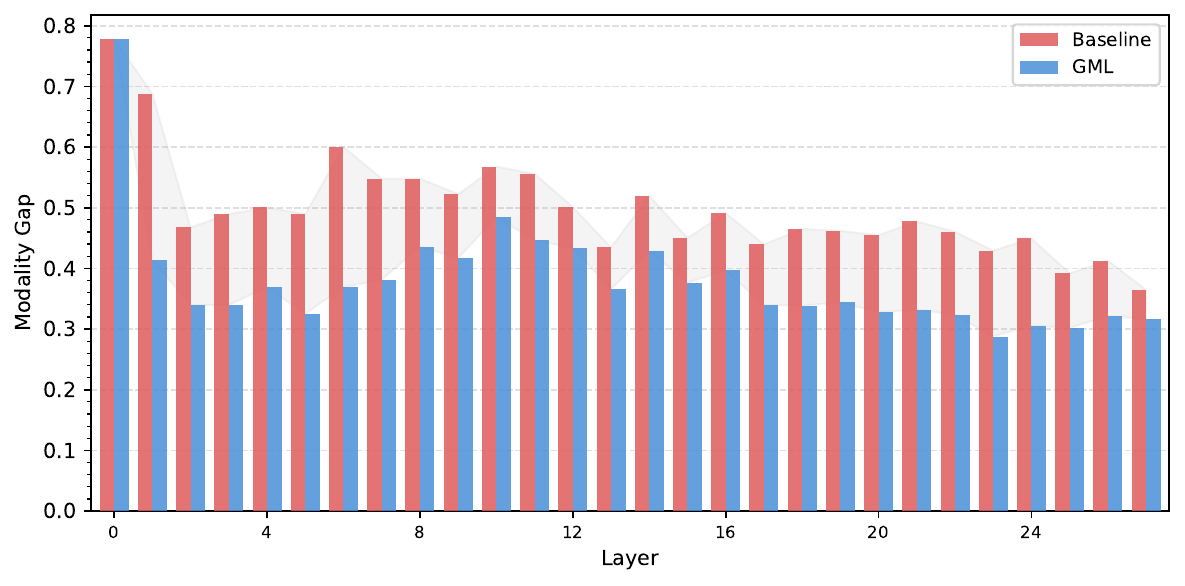} \\[6pt]
      \makebox[\linewidth]{\small\textbf{Qwen2.5-VL-7B on MMMU}} \\
      \includegraphics[width=0.92\linewidth]{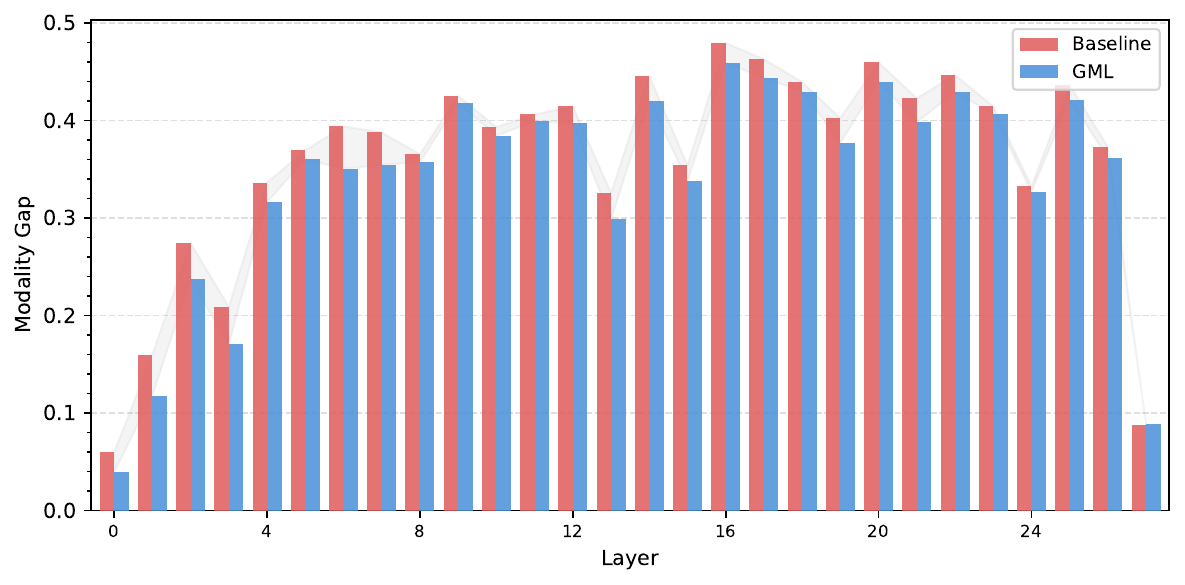} \\
    \end{tabular}
  \end{minipage}%
  }

  \caption{%
    \textbf{Left:} t-SNE visualization of hidden representations for Base and MaLoRA
    on LLaVA-1.5-7B, evaluated on MMMU (left column) and MMBench-EN (right column).
    \textbf{Right:} Representation gap analysis for
    Qwen3-VL-8B on MMBench-EN (top) and Qwen2.5-VL-7B on MMMU (bottom).%
  }
  \label{fig:tsne_gap}
\end{figure*}

Qualitatively, the base model shows a clear separation between visual and textual keys
in the t-SNE reduced space across both benchmarks (\cref{fig:tsne_gap}, left),
suggesting an inherent cross-modal key-space discrepancy.
With our method, the two distributions become noticeably closer and more overlapping,
indicating that this discrepancy is substantially alleviated.
The layer-wise modality gap analysis (\cref{fig:tsne_gap}, right) further confirms
that this separation persists across layers in the base model, while MaLoRA
consistently reduces the gap throughout the network on both Qwen3-VL-8B (MMBench-EN)
and Qwen2.5-VL-7B (MMMU), suggesting that visual keys are progressively better
integrated into the textual key distribution.

This key space separation affects attention computation. From Eq.~(\ref{eq:attn}), when $\mathbf{q}$ is better aligned with the textual key distribution, $\mathbf{q}^\top \mathbf{k}^{\text{text}}$ tends to be larger, while $\mathbf{q}^\top \mathbf{k}^{\text{vis}}$ tends to be smaller and is further down-weighted after softmax.
Consequently, even when visual tokens contain question-critical information, the model is more likely to allocate attention to text tokens or previously generated context, resulting in insufficient use of image evidence when generating the answer.
These results suggest that visual–textual key misalignment is an important factor in \emph{text-centric bias}. 

\begin{table*}[ht]
\centering
\small
\caption{%
Key-space divergence at decoder layer 1 for LLaVA-1.5-7B, Qwen2.5-VL-7B, and Qwen3-VL-8B on MMBench-EN and MMMU (10-option).
MMD and JS show Image\,vs.\,Text consistently larger than Image\,vs.\,Image and Text\,vs.\,Text.
This gap aligns with a persistent separation between visual and textual keys in the projected embedding space.%
}
\label{tab:divergence_keyspace}

\begin{subtable}[t]{0.49\linewidth}
\centering
\caption{MMD}
\resizebox{\linewidth}{!}{%
\begin{tabular}{@{}l|cc|cc|cc@{}}
\toprule
& \multicolumn{2}{c|}{LLaVA-1.5-7B}
& \multicolumn{2}{c|}{Qwen2.5-VL-7B}
& \multicolumn{2}{c}{Qwen3-VL-8B} \\
\cmidrule(lr){2-3}\cmidrule(lr){4-5}\cmidrule(lr){6-7}
Comparison & MMB-EN & MMMU & MMB-EN & MMMU & MMB-EN & MMMU \\
\midrule
\rowcolor{black!8}
Image\,vs.\,Text   & \textbf{0.94} & \textbf{0.95} & \textbf{0.57} & \textbf{0.71} & \textbf{0.76} & \textbf{0.76} \\
Image\,vs.\,Image  & 0.01 & 0.01 & 0.01 & 0.01 & 0.01 & 0.09 \\
Text\,vs.\,Text    & 0.02 & 0.02 & 0.01 & 0.07 & 0.04 & 0.03 \\
\bottomrule
\end{tabular}}
\end{subtable}
\hfill
\begin{subtable}[t]{0.49\linewidth}
\centering
\caption{JS}
\resizebox{\linewidth}{!}{%
\begin{tabular}{@{}l|cc|cc|cc@{}}
\toprule
& \multicolumn{2}{c|}{LLaVA-1.5-7B}
& \multicolumn{2}{c|}{Qwen2.5-VL-7B}
& \multicolumn{2}{c}{Qwen3-VL-8B} \\
\cmidrule(lr){2-3}\cmidrule(lr){4-5}\cmidrule(lr){6-7}
Comparison & MMB-EN & MMMU & MMB-EN & MMMU & MMB-EN & MMMU \\
\midrule
\rowcolor{black!8}
Image\,vs.\,Text   & \textbf{0.82} & \textbf{0.84} & \textbf{0.86} & \textbf{0.86} & \textbf{0.72} & \textbf{0.66} \\
Image\,vs.\,Image  & 0.04 & 0.03 & 0.04 & 0.04 & 0.04 & 0.09 \\
Text\,vs.\,Text    & 0.14 & 0.10 & 0.14 & 0.14 & 0.15 & 0.15 \\
\bottomrule
\end{tabular}}
\end{subtable}
\end{table*}

\section{Method}
\label{sec:method}

To mitigate \emph{text-centric bias} in MLLMs under joint image-text inputs, we propose the \textbf{MaLoRA} framework. It consists of two core components:
1) \textbf{Architecture}: \textbf{GML} adds modality-specific LoRA branches to the decoder self-attention $k_{proj}$ and uses a lightweight gate to enable modality-conditioned key adaptation.
2) \textbf{Regularization}: we add training-time constraints, including MK-MMD for distribution alignment and Gram-matrix regularization for structure preservation, to encourage a modality-balanced \emph{key space} without changing inference.

Following the self-attention formulation in Eq.~(\ref{eq:attn}), for the decoder at layer $l$, the input hidden states are $\mathbf{H}^{(l)}$. We assume that $\mathbf{K}^{(l)}_{\text{vis}}$ and $\mathbf{K}^{(l)}_{\text{text}}$ exhibit persistent misalignment in the key space, which is associated with decoder queries preferentially matching text tokens.

\subsection{Gated Modality LoRA (GML)}
\label{sec:method_gml}
To enable token-wise modality conditioning, we extend $k_{proj}$ with a gated two-branch LoRA design.
Given input $\mathbf{H}^{(l)}$, the resulting key representation is:
\begin{equation}
\mathbf{K}^{(l)} = \mathbf{H}^{(l)}\mathbf{W}^{(l)}_{k} + \alpha \cdot \left[ g \cdot \Delta\mathbf{K}^{(l)}_{\text{vis}} + (1-g) \cdot \Delta\mathbf{K}^{(l)}_{\text{text}} \right],
\end{equation}
where $\Delta\mathbf{K}_{\text{vis}}$ and $\Delta\mathbf{K}_{\text{text}}$ denote the visual and textual LoRA branches, respectively. The gate $g \in [0, 1]$ is predicted by a lightweight gating network:
\begin{equation}
g = \sigma\left( \text{MLP}(\mathbf{H}^{(l)}) \right),
\end{equation}
where $\sigma$ is the sigmoid activation.
To ensure precise modality-specific routing, we supervise the gating network with a token-level binary cross-entropy loss $\mathcal{L}_{\text{gate}}$ using known modality labels from the input packing:

\begin{equation}
\mathcal{L}_{\text{gate}}^{(l)} = -\left[ y \log(g) + (1-y) \log(1-g) \right],
\end{equation}
where $y$ is the modality ground truth for the current token. During training, we adopt a linear annealing schedule that transitions the gating mechanism from early label guidance to fully model-predicted gating.

\subsection{Distribution Alignment via Multi-kernel MMD}
\label{sec:method_mmd}
MaLoRA uses multi-kernel maximum mean discrepancy (MK-MMD) to reduce the distribution gap between visual and textual keys. For layer $l$, we define the alignment loss as:
\begin{equation}
\mathcal{L}^{(l)}_{\text{mmd}} = \mathrm{MMD}^{2}\!\left(\mathcal{K}^{(l)}_{\text{vis}},\mathcal{K}^{(l)}_{\text{text}}\right).
\end{equation}
Given a kernel family $\{k_{m}\}$, the multi-kernel form is $k(\mathbf{a},\mathbf{b})=\sum \alpha_{m}k_{m}(\mathbf{a},\mathbf{b})$. 
Minimizing this term reduces cross-modal distribution mismatch in the key space

\subsection{Structure Preservation via Gram Reference Regularization}
To prevent alignment from inducing excessive geometric drift, we introduce a Gram-based regularizer that preserves the relative geometry of the original key representations.
Let $\mathbf{A}^{(l)}_{\theta}$ denote the set of visual keys at layer $l$ in the current model. Its Gram matrix is defined as $\mathbf{G}(\mathbf{A})=\mathbf{A}\mathbf{A}^{\top}$. The structure preservation loss is:
\begin{equation}
\mathcal{L}^{(l)}_{\text{gram}} = \left\| \mathbf{G}\!\left(\mathbf{A}^{(l)}_{\theta}\right) - \mathbf{G}\!\left(\mathbf{A}^{(l)}_{\theta_{\text{ref}}}\right) \right\|_{F}^{2}.
\end{equation}
This constraint preserves correlation structure in a second-order sense, helping that the model retains its original reasoning capability while mitigating bias.

\subsection{Overall Objective}
\label{sec:method_objective}
The final training objective $\mathcal{L}$ combines the task loss, distribution alignment loss, structure preservation loss, and gate supervision loss:
\begin{equation}
\begin{split}
\mathcal{L}
= \mathcal{L}_{\text{task}}
+ \lambda_{\text{mmd}}\sum_{l\in\mathcal{L}}\mathcal{L}^{(l)}_{\text{mmd}}
+ \lambda_{\text{gram}}\sum_{l\in\mathcal{L}}\mathcal{L}^{(l)}_{\text{gram}} \\
\quad
+ \lambda_{\text{gate}}\sum_{l\in\mathcal{L}}\mathcal{L}^{(l)}_{\text{gate}} .
\end{split}
\end{equation}

Here $\lambda_{\text{mmd}}$, $\lambda_{\text{gram}}$, and $\lambda_{\text{gate}}$ are hyperparameters. At inference time, GML predicts token-wise gates on the fly, applying modality-conditioned key adaptation without changing the decoding procedure.

\begin{table*}[t]
\centering
\small
\caption{Per-benchmark results on three instruct models (LLaVA-1.5-7B, Qwen2.5-VL-7B, and Qwen3-VL-8B) under four training settings (Base, LoRA, QLoRA, and Ours), where Base denotes no task fine-tuning.
\textbf{Data} reports the number of training examples used for fine-tuning on each benchmark.
Benchmarks are grouped by capability categories, covering a broad and diverse evaluation suite.
For each dataset and model, the best result among the four settings is shown in bold.}
\label{tab:main_results_per_benchmark}
\resizebox{\linewidth}{!}{%
\begin{tabular}{@{}l l c | c c c c | c c c c | c c c c @{}}
\toprule
\multirow{2}{*}{Category} & \multirow{2}{*}{Benchmark} & \multirow{2}{*}{Data}
& \multicolumn{4}{c|}{LLaVA-1.5-7B}
& \multicolumn{4}{c|}{Qwen2.5-VL-7B}
& \multicolumn{4}{c}{Qwen3-VL-8B} \\
\cmidrule(lr){4-7}\cmidrule(lr){8-11}\cmidrule(lr){12-15}
& & & Base & LoRA & QLoRA & Ours
      & Base & LoRA & QLoRA & Ours
      & Base & LoRA & QLoRA & Ours \\
\midrule
\multirow{1}{*}{Gen.Understand} & MMBench-EN & 3.5k
& $72.52_{\pm 0.23}$ & $80.25_{\pm 0.35}$ & $80.95_{\pm 0.00}$ & $\mathbf{82.80_{\pm 0.35}}$
& $90.30_{\pm 0.27}$ & $90.18_{\pm 0.24}$ & $88.91_{\pm 0.07}$ & $\mathbf{92.84_{\pm 0.24}}$
& $91.34_{\pm 0.40}$ & $92.03_{\pm 0.12}$ & $89.03_{\pm 0.20}$ & $\mathbf{93.53_{\pm 0.29}}$ \\

\midrule
\multirow{2}{*}{Expert Reasoning}
& SimpleVQA  & 1.8k
& $14.16_{\pm 0.00}$ & $18.20_{\pm 0.58}$ & $19.32_{\pm 0.44}$ & $\mathbf{19.78_{\pm 0.46}}$
& $35.51_{\pm 0.13}$ & $37.98_{\pm 0.51}$ & $35.51_{\pm 0.55}$ & $\mathbf{39.33_{\pm 0.25}}$
& $37.98_{\pm 0.71}$ & $38.65_{\pm 0.76}$ & $36.63_{\pm 0.76}$ & $\mathbf{39.78_{\pm 0.38}}$ \\
& MMStar     & 1.2k
& $36.67_{\pm 1.15}$ & $36.67_{\pm 1.02}$ & $37.33_{\pm 0.88}$ & $\mathbf{37.35_{\pm 0.51}}$
& $65.67_{\pm 0.67}$ & $61.67_{\pm 0.88}$ & $59.33_{\pm 0.69}$ & $\mathbf{66.67_{\pm 0.58}}$
& $67.67_{\pm 0.19}$ & $70.33_{\pm 0.38}$ & $67.00_{\pm 0.51}$ & $\mathbf{75.00_{\pm 0.67}}$ \\

\midrule
\multirow{3}{*}{Math Reasoning}
& WeMath  & 5.8k
& $31.49_{\pm 0.15}$ & $35.00_{\pm 0.17}$ & $35.92_{\pm 0.18}$ & $\mathbf{36.84_{\pm 0.13}}$
& $58.33_{\pm 0.10}$ & $62.87_{\pm 0.11}$ & $57.30_{\pm 0.09}$ & $\mathbf{63.97_{\pm 0.09}}$
& $58.47_{\pm 0.00}$ & $68.10_{\pm 0.03}$ & $59.19_{\pm 0.13}$ & $\mathbf{70.75_{\pm 0.07}}$ \\
& MathVision & 2.8k
& $17.04_{\pm 0.09}$ & $16.74_{\pm 0.00}$ & $16.89_{\pm 0.40}$ & $\mathbf{17.04_{\pm 0.38}}$
& $23.62_{\pm 0.31}$ & $23.77_{\pm 0.30}$ & $22.42_{\pm 0.23}$ & $\mathbf{27.35_{\pm 0.52}}$
& $27.20_{\pm 0.30}$ & $27.35_{\pm 0.43}$ & $23.92_{\pm 0.17}$ & $\mathbf{30.19_{\pm 0.26}}$ \\
& DynaMath  & 4.0k
& $14.47_{\pm 0.00}$ & $26.65_{\pm 0.30}$ & $28.44_{\pm 0.06}$ & $\mathbf{28.54_{\pm 0.17}}$
& $36.92_{\pm 0.06}$ & $35.33_{\pm 0.25}$ & $36.72_{\pm 0.20}$ & $\mathbf{38.52_{\pm 0.06}}$
& $33.63_{\pm 0.17}$ & $36.23_{\pm 0.17}$ & $34.13_{\pm 0.23}$ & $\mathbf{41.42_{\pm 0.23}}$ \\

\midrule
\multirow{3}{*}{OCR QA}
& OCRVQA        & 801.6k
& $60.99_{\pm 0.01}$ & $65.96_{\pm 0.01}$ & $65.62_{\pm 0.01}$ & $\mathbf{67.33_{\pm 0.01}}$
& $79.73_{\pm 0.02}$ & $81.28_{\pm 0.00}$ & $72.98_{\pm 0.01}$ & $\mathbf{81.46_{\pm 0.02}}$
& $76.69_{\pm 0.01}$ & $81.02_{\pm 0.01}$ & $77.37_{\pm 0.02}$ & $\mathbf{81.39_{\pm 0.01}}$ \\
& TextVQA       & 34.6k
& $47.80_{\pm 0.00}$ & $50.24_{\pm 0.05}$ & $50.00_{\pm 0.04}$ & $\mathbf{50.56_{\pm 0.04}}$
& $71.14_{\pm 0.04}$ & $73.30_{\pm 0.02}$ & $71.26_{\pm 0.02}$ & $\mathbf{73.34_{\pm 0.06}}$
& $72.20_{\pm 0.01}$ & $72.94_{\pm 0.05}$ & $71.52_{\pm 0.05}$ & $\mathbf{73.28_{\pm 0.02}}$ \\
& ST-VQA        & 20.9k
& $52.16_{\pm 0.03}$ & $54.99_{\pm 0.05}$ & $54.59_{\pm 0.05}$ & $\mathbf{55.74_{\pm 0.04}}$
& $82.03_{\pm 0.06}$ & $84.23_{\pm 0.04}$ & $80.86_{\pm 0.03}$ & $\mathbf{84.37_{\pm 0.04}}$
& $81.27_{\pm 0.04}$ & $82.72_{\pm 0.01}$ & $80.19_{\pm 0.02}$ & $\mathbf{82.86_{\pm 0.02}}$ \\

\midrule
\multirow{2}{*}{Structured QA}
& DocVQA          & 39.5k
& $21.82_{\pm 0.05}$ & $24.45_{\pm 0.04}$ & $23.70_{\pm 0.04}$ & $\mathbf{25.37_{\pm 0.04}}$
& $78.24_{\pm 0.02}$ & $91.64_{\pm 0.04}$ & $90.30_{\pm 0.06}$ & $\mathbf{91.83_{\pm 0.01}}$
& $92.67_{\pm 0.05}$ & $92.54_{\pm 0.04}$ & $92.05_{\pm 0.01}$ & $\mathbf{92.88_{\pm 0.05}}$ \\
& ChartQA & 28.3k
& $14.36_{\pm 0.12}$ & $19.00_{\pm 0.12}$ & $18.12_{\pm 0.07}$ & $\mathbf{20.28_{\pm 0.12}}$
& $71.84_{\pm 0.00}$ & $80.72_{\pm 0.02}$ & $71.76_{\pm 0.12}$ & $\mathbf{80.84_{\pm 0.08}}$
& $79.04_{\pm 0.02}$ & $79.88_{\pm 0.08}$ & $77.88_{\pm 0.09}$ & $\mathbf{80.24_{\pm 0.07}}$ \\

\midrule
\multirow{1}{*}{GUI Grounding}
& RICO-ScreenQA   & 69.0k
& $32.23_{\pm 0.02}$ & $41.59_{\pm 0.03}$ & $39.92_{\pm 0.02}$ & $\mathbf{42.93_{\pm 0.01}}$
& $82.01_{\pm 0.01}$ & $86.18_{\pm 0.01}$ & $80.43_{\pm 0.02}$ & $\mathbf{87.31_{\pm 0.03}}$
& $82.18_{\pm 0.03}$ & $89.36_{\pm 0.01}$ & $81.01_{\pm 0.02}$ & $\mathbf{89.59_{\pm 0.03}}$ \\

\midrule
\multirow{2}{*}{Domain-Specific}
& RSVQA & 16.1k
& $50.90_{\pm 0.26}$ & $89.10_{\pm 0.32}$ & $88.40_{\pm 0.23}$ & $\mathbf{89.90_{\pm 0.31}}$
& $62.30_{\pm 0.29}$ & $87.70_{\pm 0.06}$ & $63.50_{\pm 0.26}$ & $\mathbf{87.90_{\pm 0.26}}$
& $61.00_{\pm 0.23}$ & $86.80_{\pm 0.25}$ & $58.90_{\pm 0.23}$ & $\mathbf{88.30_{\pm 0.26}}$ \\
& VQA-RAD    & 1.8k
& $37.25_{\pm 0.26}$ & $51.22_{\pm 0.59}$ & $50.11_{\pm 0.34}$ & $\mathbf{51.66_{\pm 0.59}}$
& $59.87_{\pm 0.13}$ & $62.75_{\pm 0.22}$ & $59.42_{\pm 0.26}$ & $\mathbf{63.64_{\pm 0.22}}$
& $57.64_{\pm 0.46}$ & $59.42_{\pm 0.26}$ & $56.32_{\pm 0.64}$ & $\mathbf{62.30_{\pm 0.38}}$ \\
\bottomrule
\end{tabular}}
\end{table*}

\begin{table}[t]
\centering
\small
\caption{Ablation on the WeMath benchmark with LLaVA-1.5-7B and Qwen3-VL-8B. We evaluate all combinations of MaLoRA: $\mathcal{L}_{\text{mmd}}$, $\mathcal{L}_{\text{gram}}$, and $\mathcal{L}_{\text{gate}}$.}
\label{tab:ablation_x_llava_qwen3}

\resizebox{\linewidth}{!}{%
\begin{tabular}{@{}>{\centering\arraybackslash}p{0.06\linewidth}
                >{\centering\arraybackslash}p{0.06\linewidth}
                >{\centering\arraybackslash}p{0.06\linewidth}
                |c|c@{}}
\toprule
\multicolumn{3}{c|}{Components} & \multicolumn{2}{c}{Model} \\
\cmidrule(lr){1-3}\cmidrule(lr){4-5}
$\mathcal{L}_{\text{mmd}}$ & $\mathcal{L}_{\text{gram}}$ & $\mathcal{L}_{\text{gate}}$ & LLaVA-1.5-7B & Qwen3-VL-8B \\
\midrule
           &            &            & $35.00_{\pm 0.11}$ & $68.10_{\pm 0.13}$ \\
           &            & \checkmark & $36.78_{\pm 0.03\textcolor{orange}{\uparrow 1.78}}$ & $69.88_{\pm 0.17\textcolor{orange}{\uparrow 1.78}}$ \\
           & \checkmark &            & $36.15_{\pm 0.15\textcolor{orange}{\uparrow 1.15}}$ & $68.74_{\pm 0.11\textcolor{orange}{\uparrow 0.64}}$ \\
\checkmark &            &            & $36.67_{\pm 0.13\textcolor{orange}{\uparrow 1.67}}$ & $68.16_{\pm 0.17\textcolor{orange}{\uparrow 0.06}}$ \\
           & \checkmark & \checkmark & $36.67_{\pm 0.11\textcolor{orange}{\uparrow 1.67}}$ & $69.66_{\pm 0.06\textcolor{orange}{\uparrow 1.56}}$ \\
\checkmark &            & \checkmark & $36.61_{\pm 0.15\textcolor{orange}{\uparrow 1.61}}$ & $70.00_{\pm 0.13\textcolor{orange}{\uparrow 1.90}}$ \\
\checkmark & \checkmark &            & $36.32_{\pm 0.07\textcolor{orange}{\uparrow 1.32}}$ & $68.62_{\pm 0.06\textcolor{orange}{\uparrow 0.52}}$ \\
\checkmark & \checkmark & \checkmark & $\mathbf{36.84_{\pm 0.17\textcolor{orange}{\uparrow 1.84}}}$ & $\mathbf{70.75_{\pm 0.14\textcolor{orange}{\uparrow 2.65}}}$ \\
\bottomrule
\end{tabular}%
}

\end{table}

\begin{table*}[t]
\centering
\small
\caption{%
Cross-modal key distribution alignment at the \textbf{early layer} for LLaVA-1.5-7B, Qwen2.5-VL-7B, and Qwen3-VL-8B on MMBench-EN and MMMU. 
MMD and JS divergence quantify the representation gap between image and text features in the projected key space. 
}
\label{tab:divergence_keyspace_early}

\begin{minipage}[t]{0.49\linewidth}
\centering
\resizebox{\linewidth}{!}{%
\begin{tabular}{@{}l|cc|cc|cc@{}}
\toprule
\multicolumn{7}{c}{\textbf{MMD}} \\
\midrule
& \multicolumn{2}{c|}{LLaVA-1.5-7B}
& \multicolumn{2}{c|}{Qwen2.5-VL-7B}
& \multicolumn{2}{c}{Qwen3-VL-8B} \\
\cmidrule(lr){2-3}\cmidrule(lr){4-5}\cmidrule(lr){6-7}
Setting & MMB-EN & MMMU & MMB-EN & MMMU & MMB-EN & MMMU \\
\midrule
Base     & 0.51 & 0.54 & 0.93 & 0.89 & 0.44 & 0.44 \\
Ours     & 0.44 & 0.51 & 0.82 & 0.71 & 0.41 & 0.41 \\
$\Delta$ & 0.07 & 0.03 & 0.11 & 0.18 & 0.03 & 0.03 \\
\bottomrule
\end{tabular}}
\end{minipage}
\hfill
\begin{minipage}[t]{0.49\linewidth}
\centering
\resizebox{\linewidth}{!}{%
\begin{tabular}{@{}l|cc|cc|cc@{}}
\toprule
\multicolumn{7}{c}{\textbf{JS}} \\
\midrule
& \multicolumn{2}{c|}{LLaVA-1.5-7B}
& \multicolumn{2}{c|}{Qwen2.5-VL-7B}
& \multicolumn{2}{c}{Qwen3-VL-8B} \\
\cmidrule(lr){2-3}\cmidrule(lr){4-5}\cmidrule(lr){6-7}
Setting & MMB-EN & MMMU & MMB-EN & MMMU & MMB-EN & MMMU \\
\midrule
Base     & 0.54 & 0.45 & 0.89 & 0.86 & 0.47 & 0.50 \\
Ours     & 0.37 & 0.40 & 0.79 & 0.82 & 0.40 & 0.35 \\
$\Delta$ & 0.17 & 0.05 & 0.10 & 0.04 & 0.07 & 0.15 \\
\bottomrule
\end{tabular}}
\end{minipage}

\end{table*}

\section{Experiments}
\subsection{Experimental Setup}

\paragraph{Models and Training Settings}
We evaluate MaLoRA on three representative instruction-tuned MLLMs: LLaVA-1.5-7B-Instruct, Qwen2.5-VL-7B-Instruct, and Qwen3-VL-8B-Instruct. These backbones cover different multimodal LLM families and provide a representative testbed for parameter-efficient adaptation. All models follow the standard vision-language input format, in which an image is encoded into visual tokens and concatenated with text tokens for decoding. Unless noted otherwise, we use each model's default vision encoder configuration, visual token budget, and preprocessing pipeline. We compare four training settings: Base, LoRA, QLoRA, and MaLoRA. For a fair comparison, all methods use the same training data, optimization schedule, batch size, and number of training epochs or steps. Additional implementation details are provided in the appendix.

\paragraph{Benchmarks and Task Taxonomy}
We evaluate on a diverse benchmark suite spanning seven capability groups (\cref{tab:main_results_per_benchmark}): \textbf{General Understanding} (MMBench-EN~\cite{liu2024mmbench}), \textbf{Expert Reasoning} (MMMU~\cite{yue2024mmmu}, SimpleVQA~\cite{cheng2025simplevqa}, MMStar~\cite{chen2024we}), \textbf{Math Reasoning} (WeMath~\cite{qiao2025we}, MathVision~\cite{wang2024measuring}), \textbf{OCR QA} (OCRVQA~\cite{mishra2019ocr}, TextVQA~\cite{singh2019towards}, ST-VQA~\cite{biten2019scene}), \textbf{Structured QA} (DocVQA~\cite{mathew2021docvqa}, ChartQA~\cite{masry2022chartqa}), \textbf{GUI Grounding} (RICO-ScreenQA~\cite{hsiao2025screenqa}), and \textbf{Domain-Specific} (VQA-RAD~\cite{lau2018dataset}, RSVQA~\cite{lobry2020rsvqa}). This taxonomy is intended to stress-test \emph{text-centric bias} in settings where visual evidence is essential. In \textbf{OCR}, \textbf{Structured}, and \textbf{GUI} tasks, the decisive signal comes directly from the image, such as rendered text, layout, or structural cues, so linguistic shortcuts are especially fragile. \textbf{Domain-Specific} benchmarks further introduce substantial distribution shifts, including remote sensing and radiology, which can increase reliance on textual priors.

\begin{figure}[ht]
\centering
\includegraphics[width=0.45\textwidth]{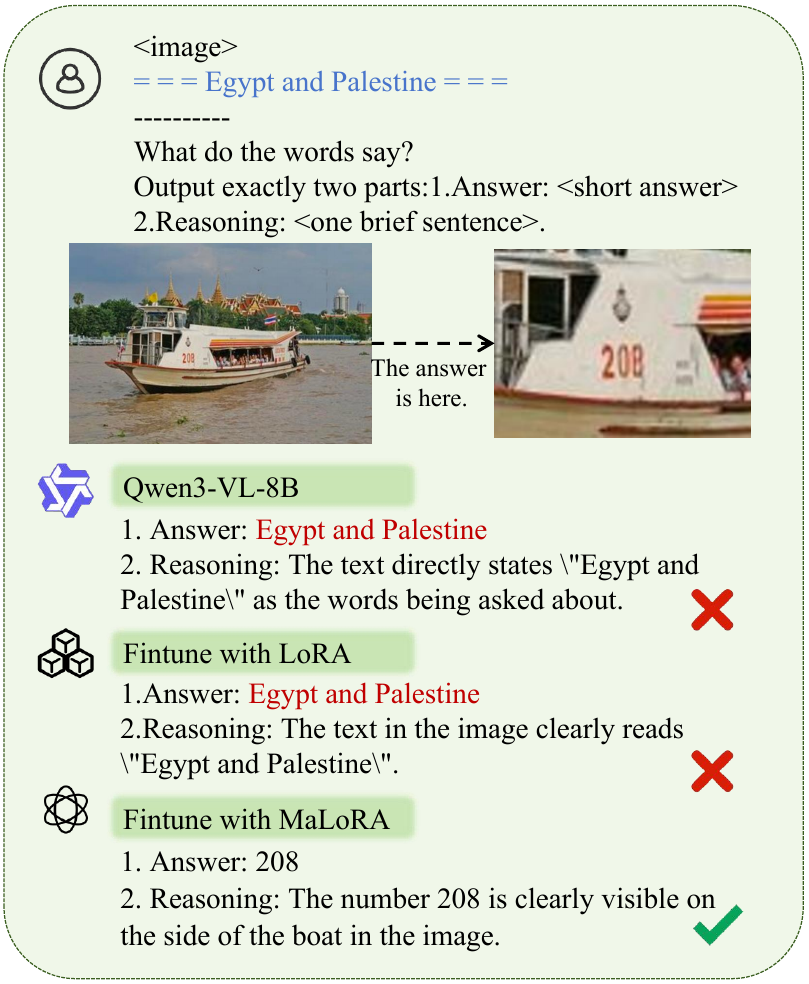}
\caption{Case study on image-text irrelevant data. MaLoRA alleviates text-centric bias, enabling the model to rely on visual evidence instead of misleading text, while LoRA and the Base model fail to do so.}
\label{fig:case_study}
\end{figure}

\subsection{Main Results}
\cref{tab:main_results_per_benchmark} reports per-benchmark results for all three backbones under four training settings: \textbf{Base} (no task fine-tuning), \textbf{LoRA}, \textbf{QLoRA}, and \textbf{Ours} (MaLoRA).

\paragraph{Overall trend}
Across 14 benchmarks covering seven capability groups, MaLoRA gives the strongest overall performance on all three backbones. It achieves the best or tied-best result on all 42 backbone-benchmark pairs. The advantage is especially clear on the two Qwen backbones, where MaLoRA ranks first on every benchmark. On LLaVA-1.5-7B, it also improves over LoRA on nearly all tasks and never drops below the best competing setting. The largest gains appear on visually demanding benchmarks such as DynaMath, MMStar, MathVision, ChartQA, and MMBench-EN, suggesting that the method is particularly effective when success depends on reliable use of image evidence rather than text priors.

To better understand this behavior, \cref{fig:case_study} shows a representative failure example from VQAv2 subsets~\cite{deng2025words}. In this example, the base model and LoRA are distracted by misleading text and fail to use the critical visual cues. MaLoRA instead grounds the prediction in the image and produces the correct answer. Additional quantitative analysis is provided in the appendix.

\paragraph{Stronger gains on vision-critical tasks}
The gains from MaLoRA are most pronounced on tasks where visual evidence is essential. The clearest improvements appear in math-heavy benchmarks, with the largest single gain observed on \textbf{DynaMath} for Qwen3-VL-8B. Similar trends also appear on other visually intensive benchmarks, including \textbf{MathVision}, \textbf{MMStar}, and \textbf{ChartQA}. By contrast, tasks that are less sensitive to cross-modal mismatch tend to show smaller but still consistent improvements. Overall, this pattern suggests that MaLoRA is most helpful when the model must resolve fine-grained visual structure, such as diagrams, charts, layouts, or spatial cues, rather than relying on language shortcuts.

\subsection{Ablation Study}
To isolate the contribution of each component in MaLoRA, we conduct a full combinatorial ablation on the WeMath benchmark using \textbf{LLaVA-1.5-7B} and \textbf{Qwen3-VL-8B} (\cref{tab:ablation_x_llava_qwen3}). Every MaLoRA variant improves over standard LoRA, and the full model consistently performs best on both backbones, confirming that the three objectives are complementary rather than redundant.

Among single-component variants, all three objectives yield consistent gains over the LoRA baseline. Combining $\mathcal{L}_{\text{gate}}$ with either $\mathcal{L}_{\text{mmd}}$ or $\mathcal{L}_{\text{gram}}$ further improves results, whereas using $\mathcal{L}_{\text{mmd}}$ and $\mathcal{L}_{\text{gram}}$ without $\mathcal{L}_{\text{gate}}$ is less effective, especially on Qwen3-VL-8B. The best performance is achieved when all three objectives are active, suggesting that reducing cross-modal discrepancy is most effective when paired with modality-aware adaptation and structural preservation in the key space.

Since the gated dual-branch architecture itself introduces additional modality-specific capacity compared to standard LoRA, a natural question is whether the observed gains stem from the alignment objectives or simply from increased expressiveness. To disentangle these two factors, we provide a parameter-matched control experiment in Appendix, along with a hyperparameter sensitivity analysis for the three loss terms.

\subsection{Cross-modal Key-space Divergence}

Our central hypothesis is that multimodal reasoning is hindered by a geometric mismatch between visual and textual representations in the projected key space. This space is particularly important because attention weights are determined by query--key similarity: if visual keys are systematically separated from textual states, then visual evidence becomes harder for textual queries to retrieve during decoding. To test this hypothesis, we measure the divergence between image and text representations in the projected key space using both MMD and Jensen--Shannon (JS) divergence.

\cref{tab:divergence_keyspace_early} reports the results at the early layer for all three backbones on MMBench-EN and MMMU, while the corresponding middle- and late-layer results are provided in Appendix \cref{tab:divergence_keyspace_mid_late}. Across all backbones and datasets, the base models consistently exhibit a clear gap between visual and textual key distributions, indicating that cross-modal misalignment already appears at shallow layers rather than emerging only in deeper processing stages.

After applying MaLoRA, both divergence measures decrease consistently across all model--dataset pairs. At the early layer, MMD is reduced by 0.03--0.18 and JS by 0.04--0.17. These reductions show that MaLoRA directly acts on the hypothesized bottleneck by bringing visual and textual keys into a more compatible geometric configuration. The effect is especially strong for Qwen2.5-VL-7B, which exhibits the largest MMD decrease on both benchmarks, while LLaVA-1.5-7B and Qwen3-VL-8B also show stable improvements under both metrics.

Importantly, this pattern is not limited to shallow layers. As shown in Appendix \cref{tab:divergence_keyspace_mid_late}, the reduction in cross-modal divergence persists through the middle and late layers. Taken together, these results support our hypothesis that key-space misalignment is a persistent property of current MLLMs and suggest that MaLoRA improves multimodal reasoning by alleviating this mismatch at its geometric source.

\begin{figure*}[t]
    \centering
    
    \begin{subfigure}{1.0\textwidth}
        \centering
        \includegraphics[width=\textwidth]{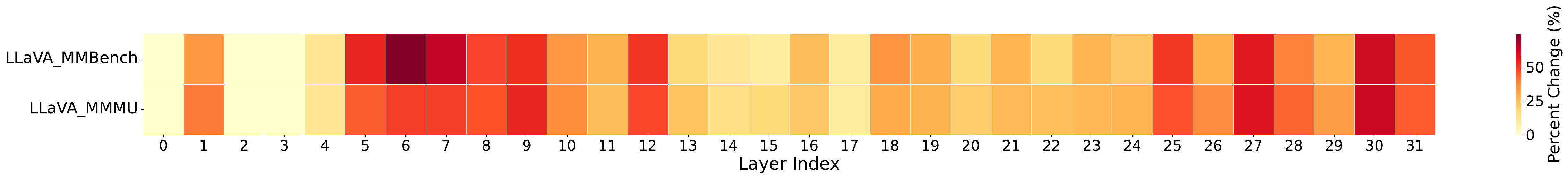}
        \label{fig:text_to_visual_llava}
    \end{subfigure}

    \caption{
Layer-wise relative change (\%) in aggregated Text$\rightarrow$Visual attention after MaLoRA fine-tuning, measured with respect to the corresponding base model across LLaVA-1.5-7B on MMMU and MMBench. Positive values indicate that textual queries allocate a larger proportion of attention to visual keys after fine-tuning.
}
    \label{fig:text_to_visual_change}
\end{figure*}

\subsection{Cross-domain Transfer}

To assess whether key-space alignment generalizes beyond the training distribution, we fine-tune each model on a single source domain (\textbf{WeMath}) and evaluate directly on nine out-of-domain benchmarks without further adaptation (\cref{tab:cross_domain}).

On \textbf{Qwen2.5-VL-7B}, MaLoRA outperforms LoRA on all nine benchmarks, improving OCR-VQA (82.29 vs.\ 80.66), ChartQA (78.28 vs.\ 78.16), DocVQA (90.26 vs.\ 89.77), SimpleVQA (39.55 vs.\ 38.20), TextVQA (72.58 vs.\ 71.66), Remote (60.80 vs.\ 60.70), ST-VQA (81.69 vs.\ 80.79), DynaMath (33.13 vs.\ 32.93), and MathVision (22.57 vs.\ 21.08). In several cases where LoRA falls substantially below the Base checkpoint, MaLoRA consistently recovers part of the performance.

For \textbf{LLaVA-1.5-7B}, MaLoRA also reduces cross-domain degradation and outperforms LoRA on all reported benchmarks, with especially large gains on \textbf{RAD} (44.57 vs.\ 37.92), \textbf{ST-VQA} (50.58 vs.\ 45.98), and \textbf{Remote} (52.90 vs.\ 51.40). Although the Base checkpoint remains strongest on some general-domain tasks, MaLoRA preserves more transferable capability than LoRA after single-domain fine-tuning. These results suggest that reducing the key-space gap during adaptation improves robustness under distribution shift.

\subsection{Attention Redistribution}

If the reduction in key-space divergence is functionally meaningful, it should also influence attention allocation by making visual tokens more accessible to textual queries. We therefore examine the layer-wise relative change in aggregated Text $\rightarrow$ Visual attention compared to the corresponding base model.

As shown in \cref{fig:text_to_visual_change}, MaLoRA increases Text $\rightarrow$ Visual attention in most layers of LLaVA-1.5-7B on both MMMU and MMBench, suggesting improved access to visual evidence during decoding. The increase is most pronounced in the early layers, where cross-modal interactions are first formed. Results for Qwen2.5-VL-7B and Qwen3-VL-8B, provided in the appendix, show a similar pattern.

Together with the divergence analysis above, these results support the proposed mechanism: reducing the geometric mismatch between visual and textual keys leads to greater attention from text tokens to visual evidence during reasoning.

\begin{table*}[t]
\centering
\small
\caption{Cross-domain transfer results. We fine-tune on WeMath2 datasets and evaluate directly on target benchmarks without further adaptation. Base uses the original instruct checkpoint with no fine-tuning. \textbf{Bold} indicates the best among all methods; \underline{underline} indicates Ours outperforms LoRA but does not surpass Base.}
\label{tab:cross_domain}
\resizebox{\textwidth}{!}{%
\begin{tabular}{@{}l|cccccccccc@{}}
\toprule
\multicolumn{11}{c}{\textbf{Qwen2.5-VL-7B} (fine-tuned on WeMath2)} \\
\midrule
Method & OCRVQA & ChartQA & DocVQA & SimpleVQA  & TextVQA & Remote & ST-VQA & DynaMath & MathVision & MMBench-EN \\
\midrule
Base   & 79.73 & 71.84 & 78.24 & 35.51 & 71.14 & 62.30 & 82.03 & 36.92 & 23.62 & 90.30  \\
LoRA   & 80.66 & 78.16 & 89.77 & 38.20 & 71.66 & 60.70 & 80.79 & 32.93 & 21.08 & 88.68\\
QLoRA  & 32.75 & 17.44 & 67.53 & 18.20 & 44.28 & 34.80 & 48.26 & 22.06 &  7.03 & 88.91 \\
Ours   & \textbf{82.29} & \textbf{78.28} & \textbf{90.26} & \textbf{39.55} & \textbf{72.58} & \underline{60.80} & \underline{81.69} & \underline{33.13} & \underline{22.57} & \underline{88.91} \\
\midrule
\multicolumn{11}{c}{\textbf{LLaVa-1.5-7B} (fine-tuned on WeMath2)} \\
\midrule
Method & OCRVQA & ChartQA & DocVQA & SimpleVQA  & TextVQA & Remote & ST-VQA & DynaMath & MathVision & MMBench-EN \\
\midrule
Base   & 60.99 & 14.36 & 21.82 & 14.16 & 47.80 & 50.90 & 52.16 & 14.47 & 17.04 & 72.52 \\
LoRA   & 58.26 & 13.08 & 20.79 & 14.61 & 46.50 & 51.40 & 45.98 & 15.46 & 12.26 & 71.36 \\
QLoRA  & 58.68 & 13.04 & 19.99 & 17.53    & 45.98 & 51.80 & 50.76    & 24.55    & 12.11 & 71.36 \\
Ours   & \underline{58.80} & \underline{13.32} & \underline{20.85} & \textbf{14.83} & \underline{46.74} & \textbf{52.90} & \underline{50.58} & \textbf{15.57} & \underline{12.86} & \underline{72.17} \\
\bottomrule
\end{tabular}}
\end{table*}

\begin{table}[t]
\centering
\small
\caption{Training and inference cost on Qwen2.5-VL-7B. Peak GPU Mem (GB) denotes maximum training memory, Step Time (s) the average time per optimization step, Params Added the number of trainable parameters, and Inference Overhead the per-sample latency relative to Base, measured over 5 GPU runs with 20 output tokens.}
\label{tab:cost_summary}
\resizebox{\linewidth}{!}{%
\begin{tabular}{@{}l|cc|cc@{}}
\toprule
Method & Peak GPU Mem (GB) & Step Time (s) & Params Added & Inference Overhead \\
\midrule
Base   & 16.87 & 237.26 & 0.00\% & $+0.00\%$ \\
LoRA   & 18.85 & 199.66 & 2.49\% & $+0.64\%$ \\
QLoRA  & 9.24  & 474.86 & 2.49\% & $+216.80\%$ \\
Ours   & 19.24 & 221.41 & 2.80\% & $+30.75\%$ \\
\bottomrule
\end{tabular}}

\end{table}

\subsection{Training and Inference Cost}

\cref{tab:cost_summary} compares MaLoRA, LoRA, and QLoRA on \textbf{Qwen2.5-VL-7B} under the same hardware and batch settings. Compared with LoRA, MaLoRA increases peak GPU memory only slightly from \textbf{18.85\,GB} to \textbf{19.24\,GB} (\textbf{+2.1\%}) and per-step time from \textbf{199.66\,s} to \textbf{221.41\,s} (\textbf{+10.9\%}). Its trainable parameters account for \textbf{2.80\%} of the base model, only marginally higher than LoRA's \textbf{2.49\%}.

At inference, LoRA adds almost no overhead after weight merging (\textbf{+0.64\%}), whereas MaLoRA incurs a \textbf{+30.75\%} latency overhead due to per-token gating in each \texttt{k\_proj} layer. However, this figure is measured with only \textbf{20 output tokens}, where the one-time prefill cost dominates. During autoregressive decoding, each step passes only a single token through a lightweight sigmoid MLP, whose $O(d)$ cost is negligible relative to attention $O(d \times L)$ and FFN $O(d \times d_{\mathrm{ff}})$, so the relative overhead diminishes as generation length grows. QLoRA shows the highest latency (\textbf{+216.80\%}) due to repeated dequantization. Overall, MaLoRA offers a practical trade-off: modest training overhead, strong accuracy gains.

\subsection{Scaling Behavior}

We further evaluate MaLoRA across four model scales of \textbf{Qwen2.5-VL} (3B, 7B, 32B, and 72B) on MMStar (\cref{tab:scaling_qwen25vl_datasetA}). MaLoRA consistently outperforms both Base and LoRA at every scale, achieving \textbf{58.67} vs.\ \textbf{57.33} (LoRA) at 3B, \textbf{66.67} vs.\ \textbf{61.67} at 7B, \textbf{71.67} vs.\ \textbf{70.00} at 32B, and \textbf{73.67} vs.\ \textbf{71.33} at 72B. The absolute gain over LoRA ranges from \textbf{+1.34} to \textbf{+5.00}, with the largest improvement at 7B. Notably, at 7B, LoRA even underperforms the Base checkpoint (61.67 vs.\ 65.67), whereas MaLoRA surpasses both. These results indicate that the benefit of key-space alignment scales well with model capacity and remains effective from small to large multimodal backbones.

\begin{table}[t]
\centering
\small
\caption{Comparison of the Base model (without task-specific fine-tuning), LoRA, and our method on MMStar across different sizes of the Qwen2.5-VL model family.}
\label{tab:scaling_qwen25vl_datasetA}
\resizebox{\linewidth}{!}{%
\begin{tabular}{@{}l|cccc@{}}
\toprule
Method & Qwen2.5-VL-3B & Qwen2.5-VL-7B & Qwen2.5-VL-32B & Qwen2.5-VL-72B \\
\midrule
Base & 55.67 & 65.67 & 65.00 & 65.33 \\
LoRA & 57.33 & 61.67 & 70.00 & 71.33 \\
Ours & 58.67 & 66.67 & 71.67 & 73.67 \\
\bottomrule
\end{tabular}}

\end{table}

\section{Conclusion}
In this paper, we show that cross-modal misalignment in the attention key space is an important source of text-centric bias in multimodal large language models. To mitigate this issue, we propose MaLoRA, a parameter-efficient fine-tuning framework that combines modality-specific key adaptation, cross-modal distribution alignment, and structure-preserving regularization. Experiments on multiple MLLM backbones and benchmarks show that MaLoRA consistently reduces the gap between visual and textual key representations and improves downstream performance, particularly on tasks that depend heavily on visual evidence. Overall, these findings suggest that aligning the attention key space is a simple and effective way to reduce modality bias in multimodal fine-tuning.

\begin{acks}
To Robert, for the bagels and explaining CMYK and color spaces.
\end{acks}

\bibliographystyle{ACM-Reference-Format}
\bibliography{sample-base}

\appendix
\onecolumn

\section{Settings}
\label{sec:settings}

\subsection{Benchmarks and Experimental Settings}

We evaluate three instruction-tuned models (LLaVA-1.5-7B, Qwen2.5-VL-7B, Qwen3-VL-8B) on public benchmarks across five core capability dimensions, covering general understanding, expert reasoning, math reasoning, Optical Character Recognition Question Answering (OCR QA), structured QA, GUI Grounding, and domain-specific tasks. The performance comparison includes four training setups (Base, LoRA, QLoRA, Ours). Detailed benchmarks used in this experiment are listed below:

\textbf{General Understanding}: One benchmark is selected to assess the models' basic vision-language understanding capabilities, namely the English version of MMBench (MMBench-EN).

\textbf{Expert Reasoning}: Two benchmarks are chosen to cover basic visual factual QA and multimodal subjective star rating evaluation, including SimpleVQA (a multimodal factuality evaluation benchmark) and MMStar (a multimodal star rating evaluation benchmark).

\textbf{Math Reasoning}: Three benchmarks focusing on visual math reasoning are adopted, covering human-level visual math, static visual math scenarios, and dynamic visual math robustness evaluation, including WeMath (a human-level visual math reasoning benchmark), MathVision (a visual math reasoning dataset), and DynaMath (a dynamic visual math reasoning robustness benchmark).

\textbf{Optical Character Recognition Question Answering (OCR QA)}: Three benchmarks are selected to evaluate OCR and text-visual QA capabilities, including OCRVQA (an OCR-based visual question answering benchmark), TextVQA (a text-based visual question answering benchmark), and ST-VQA (a scene text visual question answering benchmark).

\textbf{Structured QA}: Two benchmarks targeting structured understanding of documents and charts are selected, including DocVQA (a document image visual question answering benchmark) and ChartQA (a chart question answering reasoning benchmark).

\textbf{GUI Grounding}: One benchmark for evaluating GUI element localization capabilities is used, namely RICO-ScreenQA (a RICO screen visual question answering benchmark).

\textbf{Domain-Specific}: Two vertical domain benchmarks are chosen to cover remote sensing and medical image understanding, including RSVQA (a remote sensing benchmark) and VQA-RAD (a medical image visual question answering benchmark).

Table ~\ref{tab:lora_hparams} systematically lists the core configurations and general optimization training parameters for LoRA fine-tuning in this study. The upper section clarifies LoRA-specific hyperparameters such as rank, scaling factor, dropout probability, and target training modules, while the lower section standardizes unified training process configurations including global batch size, learning rate schedule, and precision settings. This provides a reproducible benchmark framework for LoRA and QLoRA fine-tuning of all models.

\begin{table}[H]
\centering
\small
\setlength{\tabcolsep}{6pt}
\renewcommand{\arraystretch}{1.08}

\begin{tabularx}{0.48\linewidth}{>{\raggedright\arraybackslash}X r}
\toprule
\multicolumn{2}{c}{\textbf{LoRA Fine-tuning Settings}} \\
\midrule
\textbf{Parameter} & \textbf{Value} \\
\midrule
LoRA rank ($r$) & 16 \\
LoRA scaling ($\alpha$) & 32 \\
LoRA dropout & 0.05 \\
Target modules & \texttt{q\_proj,k\_proj,v\_proj,o\_proj} \\
Bias & none \\
Trainable params & LoRA only \\
\midrule
\multicolumn{2}{c}{\textbf{Optimization \& Training}} \\
\midrule
Global batch size & 256 \\
Micro batch size & 1 \\
Gradient accumulation & 256 \\
Learning rate & 2e-4 \\
LR scheduler & cosine \\
Warmup steps & 100 \\
Weight decay & 0.0 \\
Max grad norm & 1.0 \\
Precision & bf16 \\
Max seq length & 4096 \\
Num epochs & 1 \\
\bottomrule
\end{tabularx}

\caption{Hyperparameter configuration for LoRA fine-tuning. The table is divided into two sections: LoRA-specific settings and general optimization/training settings.}
\label{tab:lora_hparams}
\end{table}

\subsection{Data Sources and Split Statistics}
To ensure transparency and reproducibility of our experimental setup, we provide dataset source descriptions, split strategies, and sample-size statistics in this section. In general, for datasets with official train/test (or train/validation/test) protocols, we strictly follow the official splits; for datasets without official splits, we randomly partition all available samples into training and test sets with an 8:2 ratio, using a fixed random seed for reproducibility.

For example, OCR-VQA follows the official train/validation/test split; MMBench-EN follows the official dev/test split; and VQA-RAD follows the official train/test split. For datasets without official split definitions, we construct training and test sets according to the same 8:2 rule.

\subsection{Grid Search Experimental Details}

To systematically explore the hyperparameter space of the three loss terms in our method, we conducted a comprehensive grid search experiment on Dataset MMStar using the Qwen3-VL-8B model. The search space covers three loss weights: $\lambda_{\text{mmd}}$ (MMD loss coefficient), $\lambda_{\text{gram}}$ (Gram matrix reference loss coefficient), and $\lambda_{\text{gate}}$ (gate supervision loss coefficient).

\textbf{Search Space Design.} We adopt a $3 \times 3 \times 3$ full factorial design, resulting in 27 unique experimental configurations. The specific search ranges are:
\begin{itemize}
    \item $\lambda_{\text{mmd}} \in \{0.05, 0.15, 0.3\}$
    \item $\lambda_{\text{gram}} \in \{0.01, 0.05, 0.1\}$
    \item $\lambda_{\text{gate}} \in \{0.05, 0.1, 0.15\}$
\end{itemize}

All experiments were completed successfully, with a total computational cost of approximately 4.5 hours (average 10.0 minutes per experiment). Table~\ref{tab:grid_search_compact} presents the complete experimental configurations in a compact factorial layout, where each cell corresponds to a specific hyperparameter combination.

\begin{table}[htbp]
\centering
\small
\setlength{\tabcolsep}{3pt}
\caption{Grid Search: $3 \times 3 \times 3$ Factorial Design. Cell values are experiment IDs.}
\label{tab:grid_search_compact}
\begin{tabular}{c|ccc|ccc|ccc}
\toprule
& \multicolumn{3}{c|}{$\lambda_{\text{gate}}=0.05$} & \multicolumn{3}{c|}{$\lambda_{\text{gate}}=0.10$} & \multicolumn{3}{c}{$\lambda_{\text{gate}}=0.15$} \\
\cmidrule(lr){2-4} \cmidrule(lr){5-7} \cmidrule(lr){8-10}
$\lambda_{\text{mmd}}$ $\backslash$ $\lambda_{\text{gram}}$ & 0.01 & 0.05 & 0.10 & 0.01 & 0.05 & 0.10 & 0.01 & 0.05 & 0.10 \\
\midrule
0.05 & 01 & 04 & 07 & 02 & 05 & 08 & 03 & 06 & 09 \\
0.15 & 10 & 13 & 16 & 11 & 14 & 17 & 12 & 15 & 18 \\
0.30 & 19 & 22 & 25 & 20 & 23 & 26 & 21 & 24 & 27 \\
\bottomrule
\end{tabular}
\end{table}
For reproducibility, Table~\ref{tab:grid_search_lambdas} provides the complete experimental metadata including experiment IDs, exact hyperparameter values, duration and evaluation scores.

\begin{table}[htbp]
\centering
\small
\caption{Grid Search Hyperparameter Configurations for Loss Weight Tuning ($3 \times 3 \times 3$ Factorial Design) with MMStar Evaluation Results}
\label{tab:grid_search_lambdas}
\setlength{\tabcolsep}{4pt}
\begin{tabular}{cccccc}
\toprule
\textbf{Exp ID} & $\lambda_{\text{mmd}}$ & $\lambda_{\text{gram}}$ & $\lambda_{\text{gate}}$ & \textbf{Train (min)} & \textbf{MMStar} \\
\midrule
01 & 0.05 & 0.01 & 0.05 & 7.5 & 73.67\% \\
02 & 0.05 & 0.01 & 0.10 & 11.0 & 73.00\% \\
03 & 0.05 & 0.01 & 0.15 & 11.9 & 72.00\% \\
04 & 0.05 & 0.05 & 0.05 & 11.2 & 72.33\% \\
05 & 0.05 & 0.05 & 0.10 & 11.4 & 72.00\% \\
06 & 0.05 & 0.05 & 0.15 & 9.5 & 72.67\% \\
07 & 0.05 & 0.10 & 0.05 & 9.7 & 73.33\% \\
08 & 0.05 & 0.10 & 0.10 & 9.6 & 73.00\% \\
09 & 0.05 & 0.10 & 0.15 & 9.3 & 72.00\% \\
10 & 0.15 & 0.01 & 0.05 & 9.7 & 73.00\% \\
11 & 0.15 & 0.01 & 0.10 & 7.9 & 74.67\% \\
12 & 0.15 & 0.01 & 0.15 & 8.9 & 72.33\% \\
13 & 0.15 & 0.05 & 0.05 & 8.7 & 73.00\% \\
14 & 0.15 & 0.05 & 0.10 & 8.3 & 73.33\% \\
15 & 0.15 & 0.05 & 0.15 & 9.1 & 72.33\% \\
16 & 0.15 & 0.10 & 0.05 & 8.8 & 72.67\% \\
17 & 0.15 & 0.10 & 0.10 & 10.2 & 72.00\% \\
18 & 0.15 & 0.10 & 0.15 & 10.2 & 74.00\% \\
19 & 0.30 & 0.01 & 0.05 & 12.3 & 71.67\% \\
20 & 0.30 & 0.01 & 0.10 & 10.3 & 73.67\% \\
21 & 0.30 & 0.01 & 0.15 & 10.3 & 73.67\% \\
22 & 0.30 & 0.05 & 0.05 & 11.0 & 73.00\% \\
23 & 0.30 & 0.05 & 0.10 & 10.4 & 73.67\% \\
24 & 0.30 & 0.05 & 0.15 & 10.2 & 73.33\% \\
25 & 0.30 & 0.10 & 0.05 & 10.1 & 73.00\% \\
26 & 0.30 & 0.10 & 0.10 & 11.2 & 72.67\% \\
27 & 0.30 & 0.10 & 0.15 & 11.1 & 75.00\% \\
\bottomrule
\end{tabular}
\end{table}

The grid search results enable us to analyze the individual and interaction effects of different loss terms on model performance. Based on these experiments, we select the optimal hyperparameter configuration for subsequent evaluations on the full benchmark suite.

\section{Additional Analysis and Experimental Results}
\subsection{Cross-modal Key-space Divergence at Middle and Late Layers}

\cref{tab:divergence_keyspace_mid_late} complements the early-layer analysis in the main text by reporting the divergence between visual and textual key distributions at the middle and late layers. Consistent with the main-text results, MaLoRA reduces cross-modal discrepancy under both MMD and Jensen--Shannon (JS) divergence across most model--dataset pairs on MMBench-EN and MMMU for all three backbones. At the middle layer, the reduction remains clearly visible, indicating that the improvement in visual--textual alignment is preserved beyond shallow representations after several layers of multimodal interaction. A similar trend is also observed at the late layer: although the absolute divergence values and the magnitude of improvement vary more across settings, MaLoRA still lowers cross-modal divergence in most cases. Overall, these results are consistent with the early-layer findings and suggest that the benefit of MaLoRA is not restricted to a single layer, but remains observable across different stages of the network.

\begin{table*}[t]
\centering
\small
\caption{%
Cross-modal key distribution alignment at the \textbf{middle} and \textbf{late} layers for LLaVA-1.5-7B, Qwen2.5-VL-7B, and Qwen3-VL-8B on MMBench-EN and MMMU. 
MMD and JS divergence quantify the representation gap between image and text features in the projected key space. 
These additional results show that the alignment improvements of our method persist beyond shallow layers and remain observable at deeper network stages.%
}
\label{tab:divergence_keyspace_mid_late}

\begin{subtable}[t]{\linewidth}
\centering
\caption{Middle layer}
\begin{minipage}[t]{0.49\linewidth}
\centering
\caption*{MMD}
\resizebox{\linewidth}{!}{%
\begin{tabular}{@{}l|cc|cc|cc@{}}
\toprule
& \multicolumn{2}{c|}{LLaVA-1.5-7B}
& \multicolumn{2}{c|}{Qwen2.5-VL-7B}
& \multicolumn{2}{c}{Qwen3-VL-8B} \\
\cmidrule(lr){2-3}\cmidrule(lr){4-5}\cmidrule(lr){6-7}
Setting & MMB-EN & MMMU & MMB-EN & MMMU & MMB-EN & MMMU \\
\midrule
Base     & 0.43 & 0.43 & 0.06 & 0.06 & 0.12 & 0.11 \\
Ours     & 0.29 & 0.33 & 0.05 & 0.05 & 0.11 & 0.10 \\
$\Delta$ & 0.14 & 0.10 & 0.01 & 0.01 & 0.01 & 0.01 \\
\bottomrule
\end{tabular}}
\end{minipage}
\hfill
\begin{minipage}[t]{0.49\linewidth}
\centering
\caption*{JS}
\resizebox{\linewidth}{!}{%
\begin{tabular}{@{}l|cc|cc|cc@{}}
\toprule
& \multicolumn{2}{c|}{LLaVA-1.5-7B}
& \multicolumn{2}{c|}{Qwen2.5-VL-7B}
& \multicolumn{2}{c}{Qwen3-VL-8B} \\
\cmidrule(lr){2-3}\cmidrule(lr){4-5}\cmidrule(lr){6-7}
Setting & MMB-EN & MMMU & MMB-EN & MMMU & MMB-EN & MMMU \\
\midrule
Base     & 0.44 & 0.43 & 0.36 & 0.24 & 0.29 & 0.29 \\
Ours     & 0.28 & 0.34 & 0.20 & 0.19 & 0.19 & 0.17 \\
$\Delta$ & 0.16 & 0.09 & 0.16 & 0.05 & 0.10 & 0.12 \\
\bottomrule
\end{tabular}}
\end{minipage}
\end{subtable}

\vspace{0.6em}

\begin{subtable}[t]{\linewidth}
\centering
\caption{Late layer}
\begin{minipage}[t]{0.49\linewidth}
\centering
\caption*{MMD}
\resizebox{\linewidth}{!}{%
\begin{tabular}{@{}l|cc|cc|cc@{}}
\toprule
& \multicolumn{2}{c|}{LLaVA-1.5-7B}
& \multicolumn{2}{c|}{Qwen2.5-VL-7B}
& \multicolumn{2}{c}{Qwen3-VL-8B} \\
\cmidrule(lr){2-3}\cmidrule(lr){4-5}\cmidrule(lr){6-7}
Setting & MMB-EN & MMMU & MMB-EN & MMMU & MMB-EN & MMMU \\
\midrule
Base     & 0.16 & 0.22 & 0.04 & 0.23 & 0.02 & 0.02 \\
Ours     & 0.07 & 0.07 & 0.03 & 0.21 & 0.01 & 0.01 \\
$\Delta$ & 0.09 & 0.15 & 0.01 & 0.02 & 0.01 & 0.01 \\
\bottomrule
\end{tabular}}
\end{minipage}
\hfill
\begin{minipage}[t]{0.49\linewidth}
\centering
\caption*{JS}
\resizebox{\linewidth}{!}{%
\begin{tabular}{@{}l|cc|cc|cc@{}}
\toprule
& \multicolumn{2}{c|}{LLaVA-1.5-7B}
& \multicolumn{2}{c|}{Qwen2.5-VL-7B}
& \multicolumn{2}{c}{Qwen3-VL-8B} \\
\cmidrule(lr){2-3}\cmidrule(lr){4-5}\cmidrule(lr){6-7}
Setting & MMB-EN & MMMU & MMB-EN & MMMU & MMB-EN & MMMU \\
\midrule
Base     & 0.39 & 0.39 & 0.31 & 0.27 & 0.47 & 0.35 \\
Ours     & 0.32 & 0.36 & 0.28 & 0.16 & 0.41 & 0.31 \\
$\Delta$ & 0.07 & 0.03 & 0.03 & 0.09 & 0.06 & 0.04 \\
\bottomrule
\end{tabular}}
\end{minipage}
\end{subtable}

\end{table*}

\subsection{Image-Text Irrelevant Experiments}

Beyond these general benchmarks, we also construct and use an image-text irrelevant test suite to evaluate model robustness under text-visual conflict conditions (i.e., when distracting text is irrelevant to visual evidence). Specifically, we adopt the image-text irrelevant subsets from \citet{deng2025words} study as test data, and unify them into the test processed format, including three subsets: vqav2\_irrelevant, docvqa\_irrelevant, and openphish\_irrelevant, with evaluation sizes of 1,000, 1,000, and 5,000, respectively. The corresponding results are reported in \cref{tab:irrelevant}.

From \cref{tab:irrelevant}, MaLoRA achieves the best performance across most settings, with the clearest gains on openphish\_irrelevant (e.g., 5.24 vs. 2.48/2.04 for Qwen2.5-VL-7B, and 3.04 vs. 1.88/0.40 for Qwen3-VL-8B), indicating stronger robustness under severe irrelevant-text interference. On vqav2\_irrelevant and docvqa\_irrelevant, improvements are smaller but still consistent, suggesting that our method improves robustness while preserving strong baseline capability on relatively easier irrelevant scenarios.

\begin{table*}[t]
\centering
\large
\setlength{\tabcolsep}{10pt}
\renewcommand{\arraystretch}{1.1}
\begin{tabular}{@{}lcccccccc@{}}
\toprule
& \multicolumn{4}{c}{\textbf{Qwen2.5-VL-7B}} & \multicolumn{4}{c}{\textbf{Qwen3-VL-8B
}} \\
\cmidrule(lr){2-5} \cmidrule(lr){6-9}
Datasets & Base & LoRA & QLoRA & MaLoRA & Base & LoRA & QLoRA & MaLoRA \\
\midrule
vqav2\_irrelevant  & 68.4 & 71 & 58&	71.8	& 70.7 &	71.8 &	66.9 &	72.2  \\
openphish\_irrelevant  & 1.6&	2.48	&2.04&	5.24	&0.91	&1.88	&0.4	&3.04  \\
docvqa\_irrelevant & 91.6&	91.2&	83.3&	91.9&	92.8&	92.6&	91.5	&93.5  \\
\bottomrule
\end{tabular}
\caption{Performance (\%) on the irrelevant benchmark subsets (\texttt{vqav2\_irrelevant}, \texttt{openphish\_irrelevant}, and \texttt{docvqa\_irrelevant}) for Qwen2.5-VL-7B and Qwen3-VL-8B under four fine-tuning settings: Base, LoRA, QLoRA, and MaLoRA.}
\label{tab:irrelevant}
\end{table*}

\begin{figure*}[t]
    \centering
    
    \begin{subfigure}{0.9\textwidth}
        \centering
        \includegraphics[width=\textwidth]{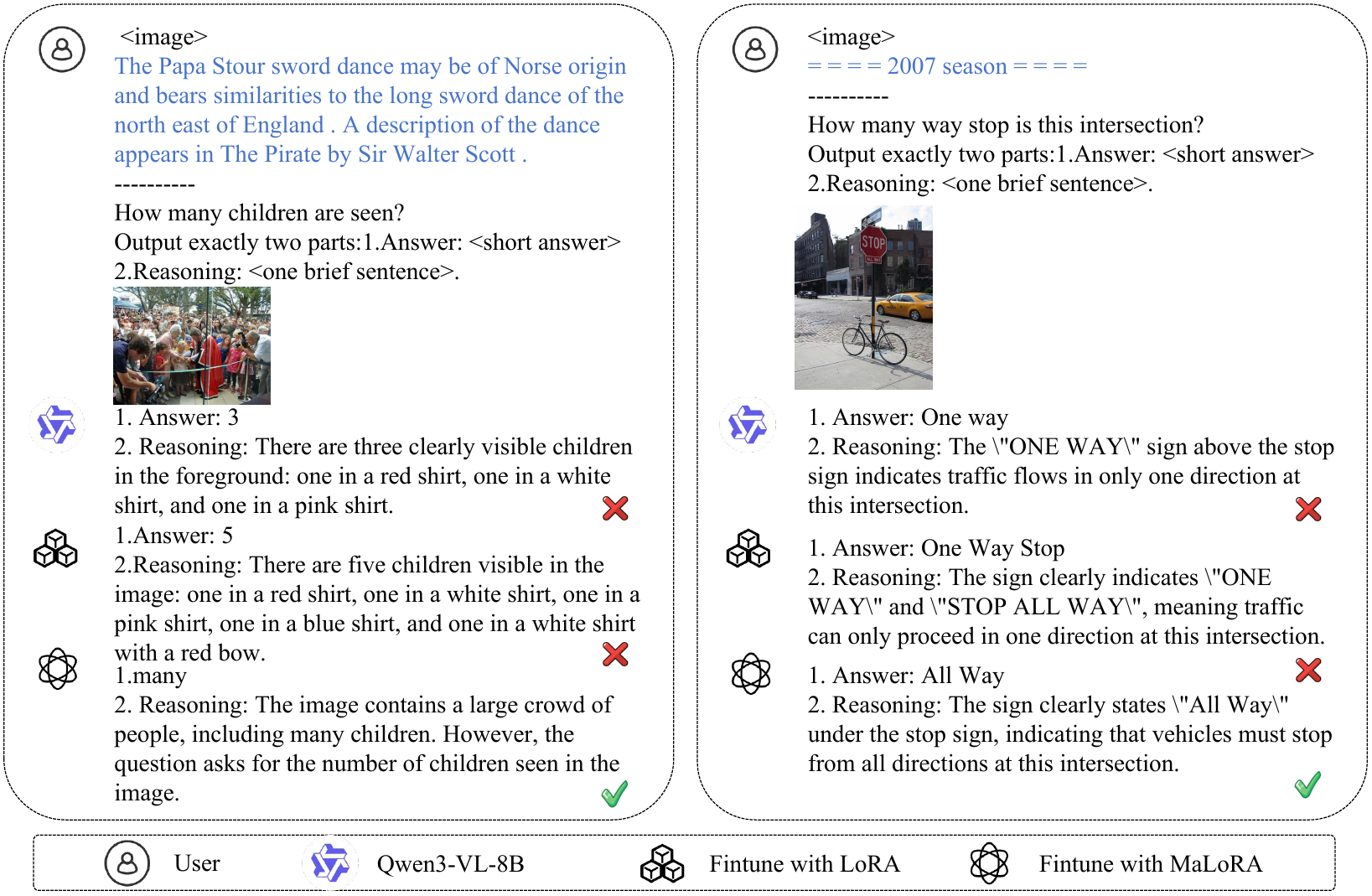}
    \end{subfigure}
    
    \caption{Case studies on the irrelevant setting. When irrelevant textual distractors are added to the prompt, the base Qwen3-VL-8B and LoRA-finetuned models are more likely to be misled, while the MaLoRA-finetuned model remains grounded in visual evidence and produces correct answers.}
    \label{fig:case_study1}
\end{figure*}

As shown in \cref{fig:case_study1}, we present two representative case studies under the irrelevant setting, where irrelevant textual distractors are inserted into the input and the model is expected to answer based on visual evidence. The results indicate that both the base Qwen3-VL-8B model and the LoRA-finetuned variant are more easily biased by textual priors, leading to errors such as incorrect counting and semantic misinterpretation of traffic signs. In contrast, the MaLoRA-finetuned model consistently ignores irrelevant text, focuses on key visual cues, and produces correct answers. These qualitative examples demonstrate that our method achieves stronger robustness and better vision-grounded alignment under text–vision conflict conditions.

\subsection{Modality Gap Metric}

We use the modality gap metric to quantify the alignment between visual and textual representations in \cref{fig:tsne_gap}. Here we provide the formal definition.

For a given transformer layer $l$, let $\{\mathbf{v}_i\}_{i=1}^{N_v} \subset \mathbb{R}^d$ and $\{\mathbf{t}_j\}_{j=1}^{N_t} \subset \mathbb{R}^d$ denote the key vectors extracted from visual tokens and text tokens, respectively. We first apply $\ell_2$ normalization to each vector:
\begin{equation}
    \hat{\mathbf{v}}_i = \frac{\mathbf{v}_i}{\|\mathbf{v}_i\|_2}, \quad
    \hat{\mathbf{t}}_j = \frac{\mathbf{t}_j}{\|\mathbf{t}_j\|_2}.
\end{equation}

The modality gap at layer $l$ is then defined as the Euclidean distance between the centroids of the two normalized distributions:
\begin{equation}
    \mathcal{G}^{(l)} = \left\| \frac{1}{N_v}\sum_{i=1}^{N_v} \hat{\mathbf{v}}_i \;-\; \frac{1}{N_t}\sum_{j=1}^{N_t} \hat{\mathbf{t}}_j \right\|_2.
\end{equation}

A smaller $\mathcal{G}^{(l)}$ indicates that the visual and textual representations are more closely aligned in the attention key space at layer $l$. Since all vectors lie on the unit hypersphere after normalization, $\mathcal{G}^{(l)} \in [0, 2]$.

\subsection{Additional t-SNE Visualization}

\begin{figure*}[t]
  \centering
  \scalebox{0.92}{%
  \begin{minipage}[c]{\textwidth}
    \centering
    \begin{tabular}{
      @{}
      c
      @{\hspace{2pt}}
      c
      @{\hspace{4pt}}
      c
      @{\hspace{4pt}}
      c
      @{\hspace{4pt}}
      c
      @{}
    }
      &
      \makebox[0.23\textwidth]{\small\textbf{Qwen3-VL-8B on MMMU}}
      &
      \makebox[0.23\textwidth]{\small\textbf{Qwen3-VL-8B on MMBench-EN}}
      &
      \makebox[0.23\textwidth]{\small\textbf{Qwen2.5-VL-7B on MMMU}}
      &
      \makebox[0.23\textwidth]{\small\textbf{Qwen2.5-VL-7B on MMBench-EN}}
      \\[2pt]

      \raisebox{-.5\height}{\rotatebox{90}{\small\textbf{Base}}}
      &
      \raisebox{-.5\height}{\includegraphics[width=0.23\textwidth]{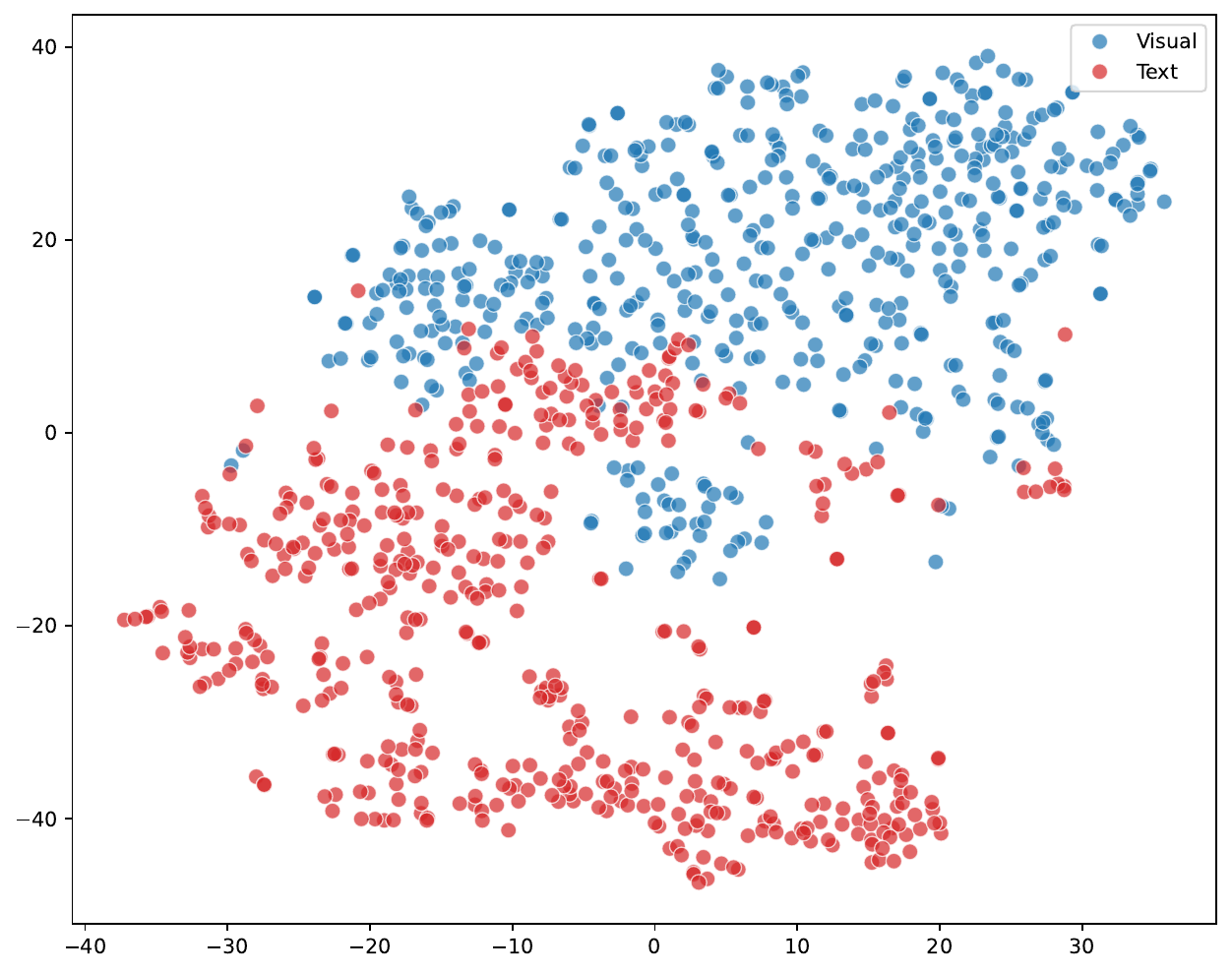}}
      &
      \raisebox{-.5\height}{\includegraphics[width=0.23\textwidth]{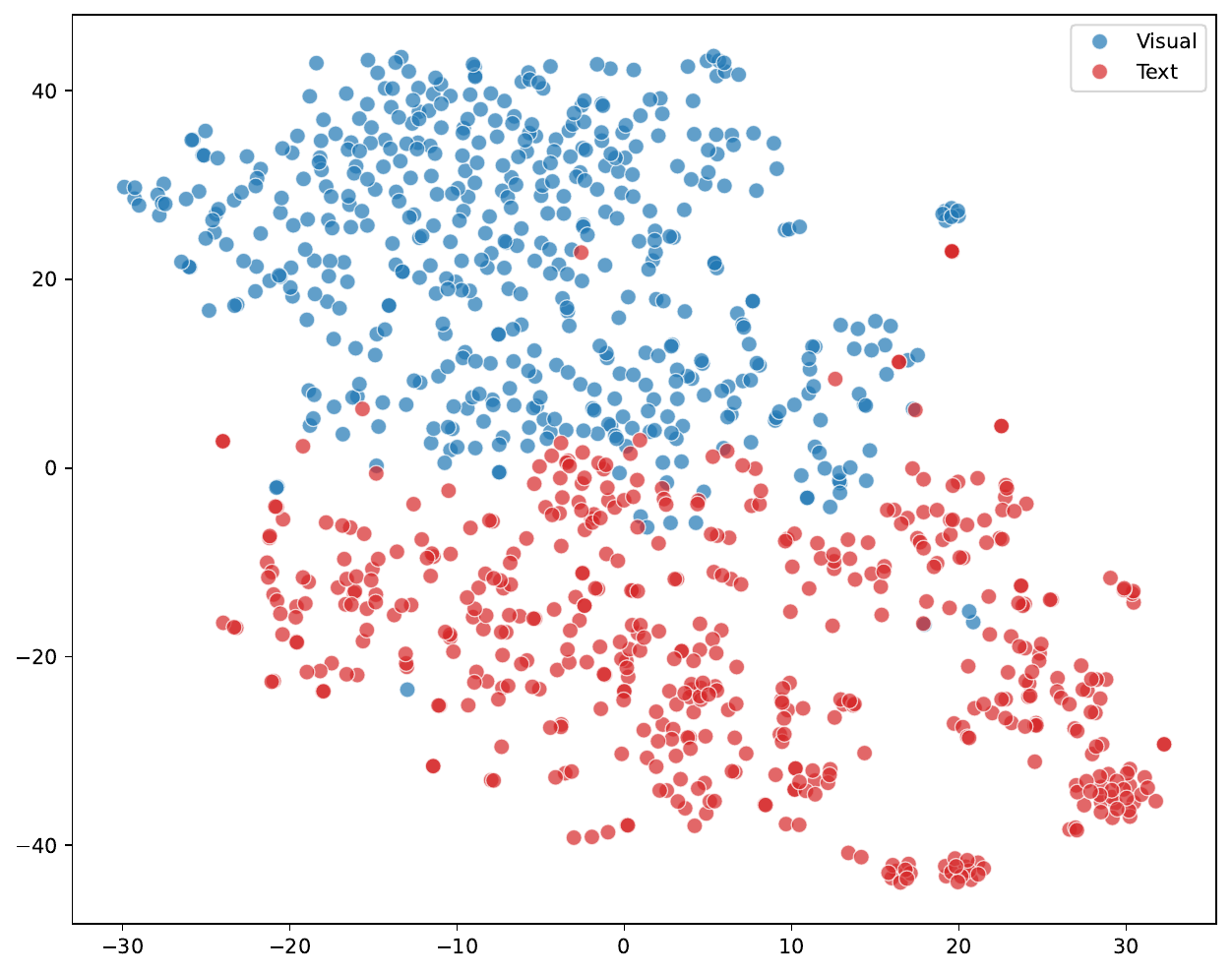}}
      &
      \raisebox{-.5\height}{\includegraphics[width=0.23\textwidth]{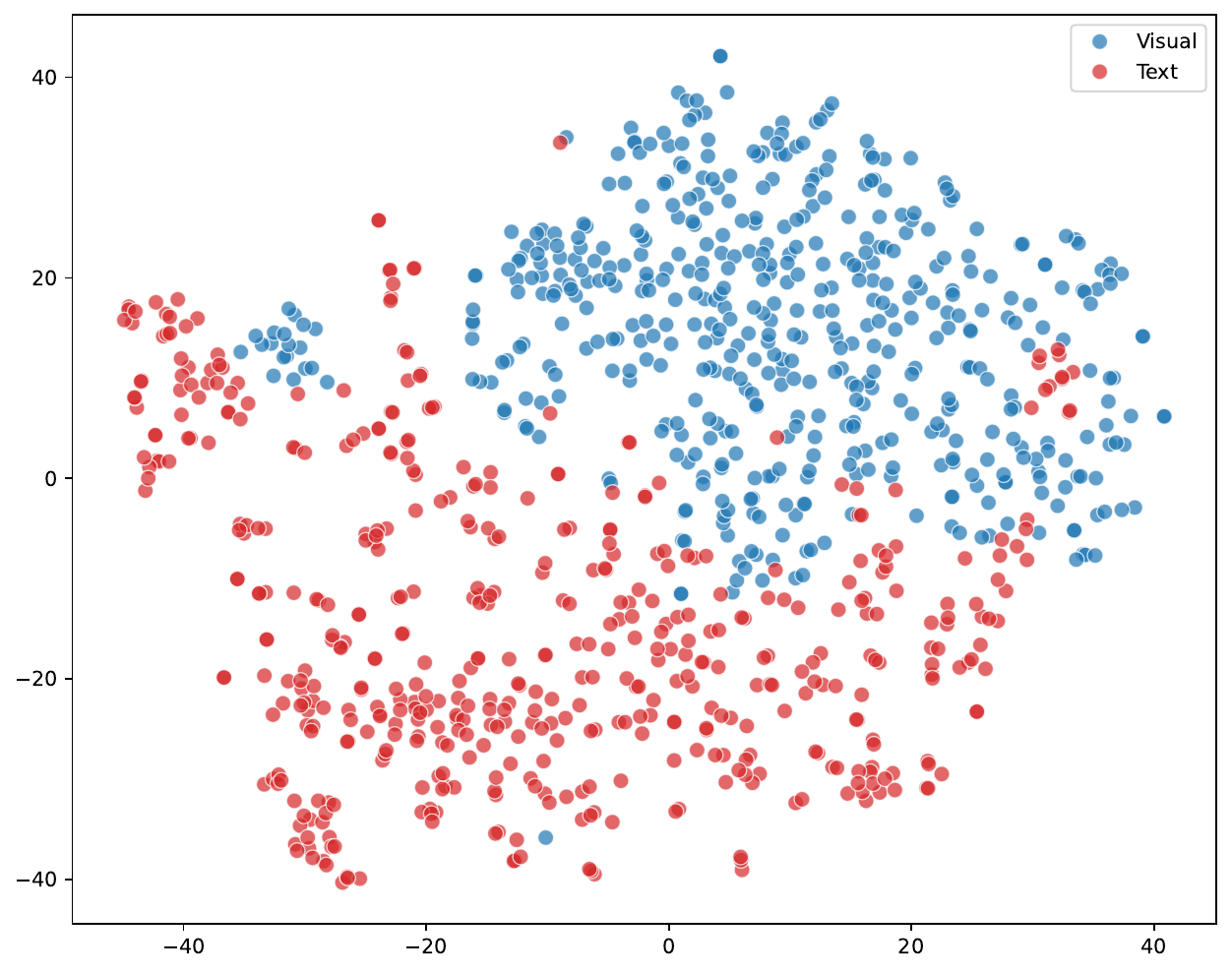}}
      &
      \raisebox{-.5\height}{\includegraphics[width=0.23\textwidth]{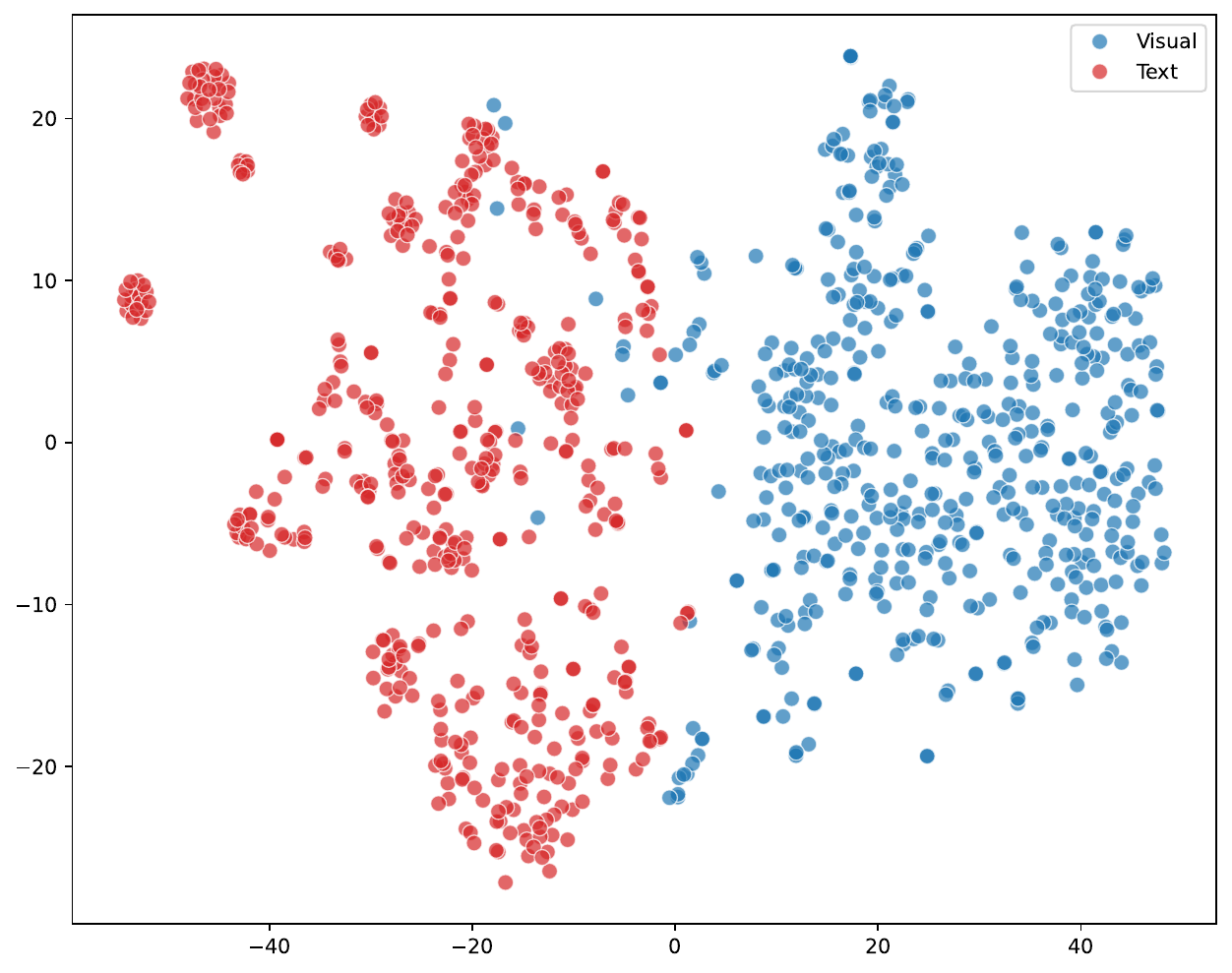}}
      \\[4pt]

      \raisebox{-.5\height}{\rotatebox{90}{\small\textbf{MaLoRA}}}
      &
      \raisebox{-.5\height}{\includegraphics[width=0.23\textwidth]{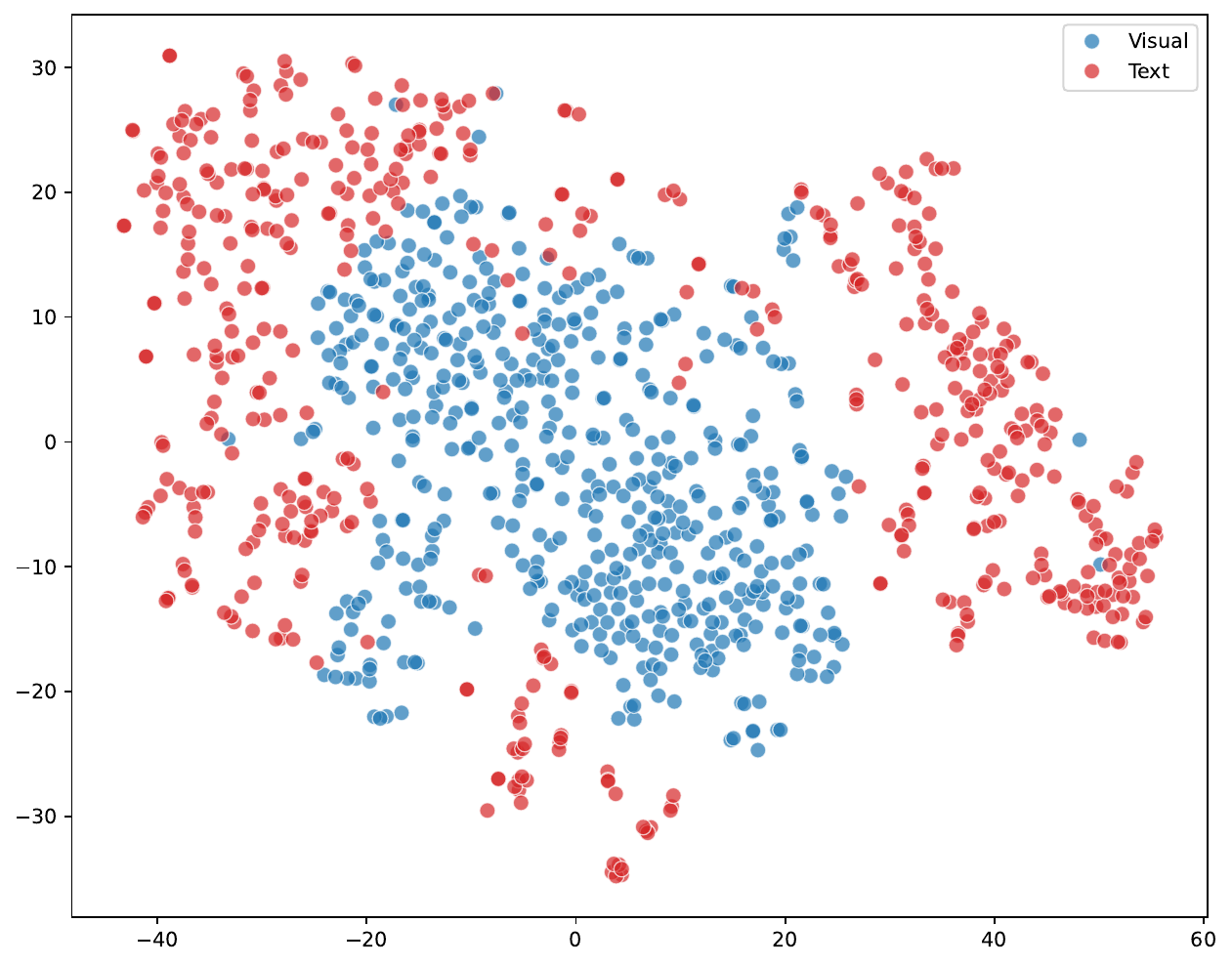}}
      &
      \raisebox{-.5\height}{\includegraphics[width=0.23\textwidth]{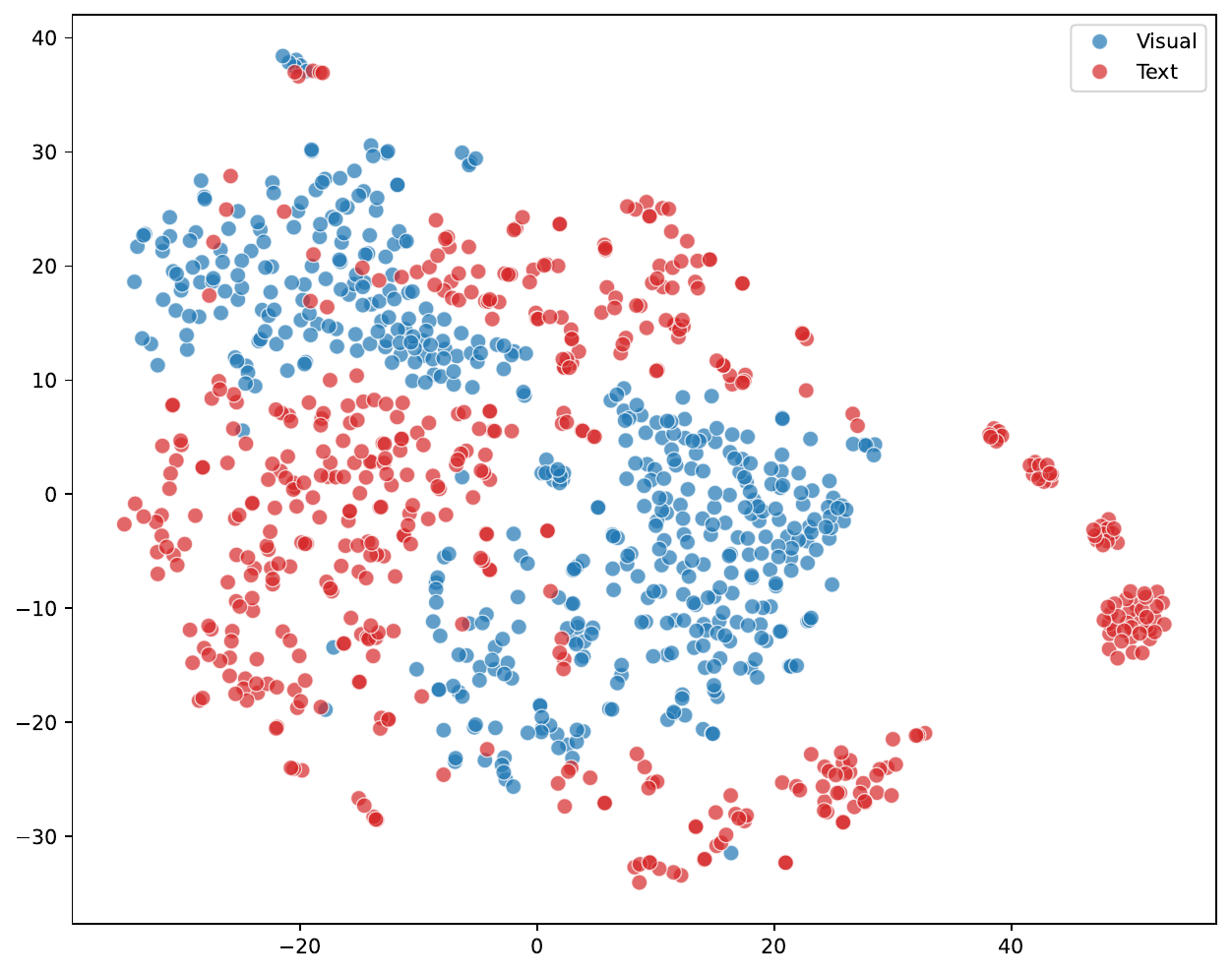}}
      &
      \raisebox{-.5\height}{\includegraphics[width=0.23\textwidth]{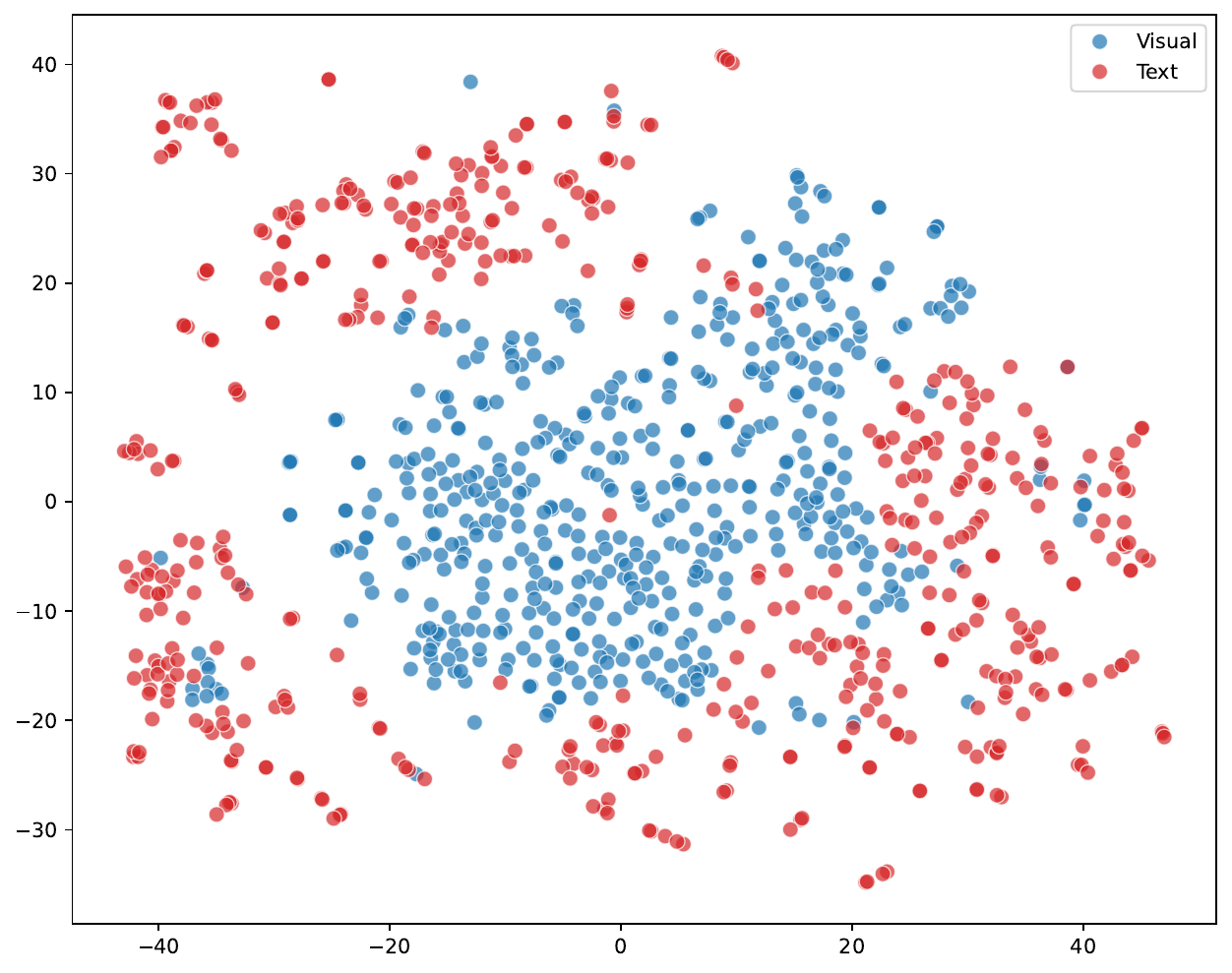}}
      &
      \raisebox{-.5\height}{\includegraphics[width=0.23\textwidth]{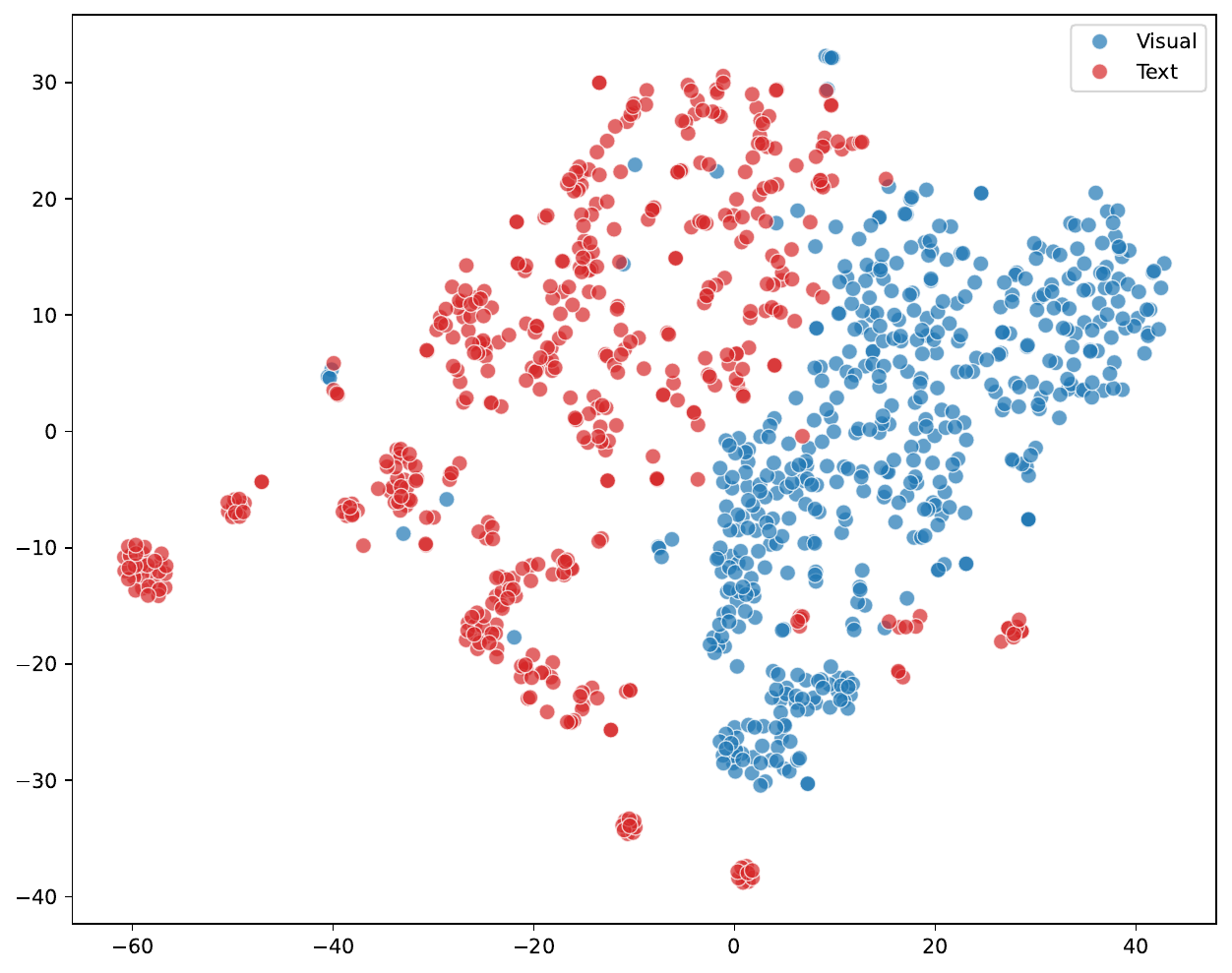}}
      \\
    \end{tabular}
  \end{minipage}%
  }

  \caption{%
    t-SNE visualization of hidden representations for Base and MaLoRA on
    Qwen3-VL-8B and Qwen2.5-VL-7B, evaluated on MMMU and MMBench-EN.
    Each column shows one model--dataset pair, and the two rows compare
    the Base model with MaLoRA.%
  }
  \label{fig:app_tsne_qwen_compare}
\end{figure*}

Figure ~\ref{fig:app_tsne_qwen_compare} presents additional t-SNE visualizations for Qwen3-VL-8B and Qwen2.5-VL-7B on MMMU and MMBench-EN. Across all four model--dataset pairs, the Base model shows a visible separation between visual and text representations, reflecting persistent cross-modal discrepancy in the key space. After fine-tuning with MaLoRA, the two modalities become noticeably closer and more mixed. This qualitative evidence further supports our main conclusion that MaLoRA consistently alleviates cross-modal key-space misalignment across different backbones and benchmarks.

\subsection{Additional Layer-wise Text-to-Visual Attention Change Analysis}
\label{app:text_to_visual_qwen}

\begin{figure*}[t]
    \centering

    \begin{subfigure}{1.0\textwidth}
        \centering
        \includegraphics[width=\textwidth]{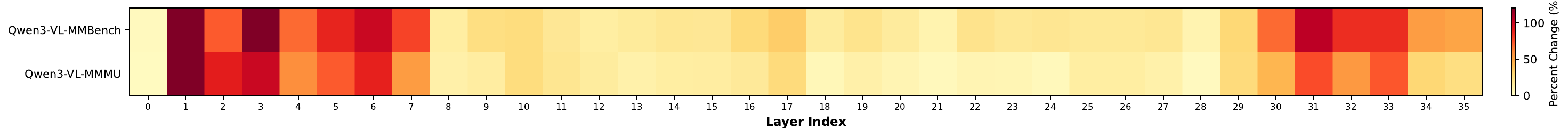}
        \caption{Qwen3-VL-8B}
        \label{fig:text_to_visual_qwen3}
    \end{subfigure}

    \vspace{6pt}

    \begin{subfigure}{1.0\textwidth}
        \centering
        \includegraphics[width=\textwidth]{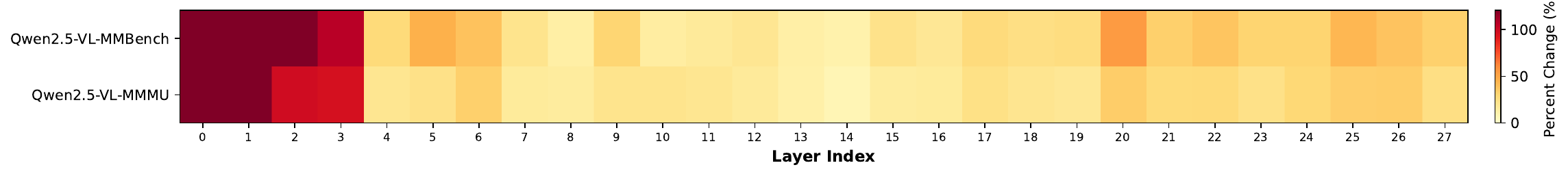}
        \caption{Qwen2.5-VL-7B}
        \label{fig:text_to_visual_qwen25}
    \end{subfigure}

    \caption{
    Layer-wise relative change (\%) in aggregated Text$\rightarrow$Visual attention after MaLoRA fine-tuning, measured with respect to the corresponding base model on Qwen3-VL-8B and Qwen2.5-VL-7B across MMMU and MMBench. Positive values indicate that textual queries allocate a larger proportion of attention to visual keys after fine-tuning.
    }
    \label{fig:text_to_visual_change_qwen}
\end{figure*}

To complement the main-text attention analysis on LLaVA-1.5-7B, \cref{fig:text_to_visual_change_qwen} further reports the layer-wise relative change in aggregated Text$\rightarrow$Visual attention for Qwen3-VL-8B and Qwen2.5-VL-7B after MaLoRA fine-tuning. These results provide additional evidence that the strengthened text-to-visual attention pattern generalizes across different MLLM backbones.

\section{Evaluation Prompts}
To ensure reproducibility and facilitate future research, we provide here the complete set of prompts used
to evaluate our model across all benchmarks. These prompts were consistently applied during inference
to maintain fairness and comparability.
\subsection{General Understanding}
\begin{algopanel}{MMBench-EN}
$<image>$

Question: {question}

Options:

{options}

Please select the correct answer from the options above.
\end{algopanel}

\subsection{Expert Reasoning}
\begin{algopanel}{MMStar}
$<image>$

Question: {question}

Options:

{options}

Please select the correct answer from the options above.
\end{algopanel}

\begin{algopanel}{SimpleVQA}
$<image>$

Question: {question}

Answer the question using a single word or phrase.
\end{algopanel}

\subsection{Math Reasoning}
\begin{algopanel}{WeMath}
$<image>$

Now, we require you to solve a multiple-choice math question. Please briefly describe your
thought process and provide the final answer(option).

Question: {question}

Option: {options}

Answer the question using a single word or phrase, strictly in the format of A, B, C, D.
\end{algopanel}

\begin{algopanel}{DynaMath $\vert$ MathVision}
$<image>$

Question: {question}

Answer the question using a single word or phrase.
\end{algopanel}

\subsection{OCR QA}
\begin{algopanel}{OCRVQA $\vert$ TextVQA $\vert$ ST-VQA}
$<image>$

Question: {question}

Answer the question using a single word or phrase.
\end{algopanel}

\subsection{Structured QA}
\begin{algopanel}{DocVQA $\vert$ ChartQA}
$<image>$

Question: {question}

Answer the question using a single word or phrase.
\end{algopanel}

\subsection{GUI Grounding}
\begin{algopanel}{RICO-ScreenQA}
$<image>$

Question: {question}

Answer the question using a single word or phrase.

\vspace{0.5cm} 
 
You are an impartial evaluator. Given a question, the ground truth answer, and a model's prediction, judge whether the prediction is semantically correct (equivalent to or appropriately conveys the ground truth).

Question: {question}

Ground Truth Answer: {answer}

Model Prediction: {prediction}

Is the model's prediction correct? Answer with ONLY "YES" or "NO".
Answer:
\end{algopanel}

\subsection{Domain-Specific}
\begin{algopanel}{RSVQA}
$<image>$

Question: {question}

Answer the question using a single word or phrase.
\end{algopanel}

\begin{algopanel}{VQA-RAD}
$<image>$

Question: {question}

Answer the question using a single word or phrase.
\end{algopanel}

\end{document}